\documentclass[letterpaper]{article} 
\usepackage{aaai24}  

\usepackage[font={small}]{caption}
\usepackage[font={scriptsize}]{subcaption}
\usepackage{paralist}
\usepackage{enumitem}
\usepackage{sidecap}
\usepackage{amssymb}
\usepackage{amsmath}
\usepackage{amsthm}
\usepackage{amsfonts}
\usepackage{dsfont}
\usepackage{bm}
\usepackage{booktabs}

\usepackage{times}  
\usepackage{helvet}  
\usepackage{courier}  
\usepackage[hyphens]{url}  
\usepackage{graphicx} 
\usepackage{natbib}  
\usepackage{caption} 
\usepackage{algorithm}
\usepackage{algorithmic}

\usepackage{color}
\usepackage{xcolor}

\definecolor{navy}{rgb}{0,0,0.0}
\definecolor{airforceblue}{rgb}{0.36, 0.54, 0.66}
\definecolor{mydarkblue}{rgb}{0,0.08,0.45}

\newtheorem*{result*}{Theorem}

\usepackage{newfloat}
\usepackage{listings}
\DeclareCaptionStyle{ruled}{labelfont=normalfont,labelsep=colon,strut=off} 
\floatstyle{ruled}
\newfloat{listing}{tb}{lst}{}
\floatname{listing}{Listing}
\title{Relearning Forgotten Knowledge: on Forgetting, Overfit and Training-Free Ensembles of DNNs}
\author{
    Uri Stern, Daphna Weinshall
}
\affiliations{
    School of Computer Science and Engineering, Hebrew University of Jerusalem, Israel\\
    ustern@gmail.com, daphna@mail.huji.ac.il\\
}

\begin{document}

\maketitle





\begin{abstract}
    The infrequent occurrence of overfit in deep neural networks is perplexing. On the one hand, theory predicts that as models get larger they should eventually become too specialized for a specific training set, with ensuing decrease in generalization. In contrast, empirical results in image classification indicate that increasing the training time of deep models or using bigger models almost never hurts generalization. Is it because the way we measure overfit is too limited? 
\hspace{0.4cm}    
    Here, we introduce a novel score for quantifying overfit, which monitors the forgetting rate of deep models on validation data. Presumably, this score indicates that even while generalization improves overall, there are certain regions of the data space where it deteriorates. When thus measured, we show that overfit can occur with and without a decrease in validation accuracy, and may be more common than previously appreciated. 
    This observation may 
    help to clarify the aforementioned confusing picture. 
\hspace{0.4cm}     
    We use our observations to construct a new ensemble method, based solely on the training history of a single network, which provides significant improvement in performance without any additional cost in training time. An extensive empirical evaluation with modern deep models shows our method's utility on multiple datasets, neural networks architectures and training schemes, both when training from scratch and when using pre-trained networks in transfer learning. Notably, our method outperforms comparable methods while being easier to implement and use, and further improves the performance of competitive networks on Imagenet by 1\%. 

\end{abstract}
\section{Introduction}

Overfit of train data constitutes a fundamental problem in machine learning. Theoretical analysis predicts that as a model acquires additional degrees of freedom, its ability to fit a certain training data increases. As a result, a model's generalization error is expected to increase when it becomes too specialized for a specific training set. Accordingly, in deep learning we expect to see \emph{increased generalization error} as the number of parameters and/or training epochs increases. Surprisingly, even vast deep neural networks with many billions of parameters rarely fulfil this expectation. In fact, even in larger models that do exhibit slightly inferior performance \citep{liu2022convnet} with increased size, we do not see overfit as a function of epochs. More typically, even substantial increase in the number of parameters will still lead to improved performance, or to more bizarre behaviors like the double descent in test error \citep[][see Section~\ref{sec:overfitanddoubledescent}]{annavarapu2021deep}. Clearly there is a gap between our classical understanding of overfit and the empirical results obtained when using modern neural networks. 

To bridge this gap, we introduce a novel perspective on overfit. Instead of gauging it solely through a reduction in \emph{test accuracy}, we propose to track what we term \emph{the model's forget fraction}. This metric represents the portion of test data\footnote{We use the common terminology, though for estimation purposes, the various scores are evaluated on validation data.} that the model initially classifies correctly but misclassifies as training proceeds. In Section~\ref{sec:overfitanddoubledescent} we examine various benchmark datasets, where we measure this phenomenon even if no overfit is observed using the traditional definition, namely, when test accuracy increases as learning proceeds. Interestingly, we observe that models can suffer from a significant level of forgetting, which indicates that our score measures something new. Importantly, this occurs even after the deployment of modern methods to reduce overfit, such as data augmentation, while using competitive networks. 

Using this new perspective with its corresponding scores and to better understand the phenomenon, we analyze in Section~\ref{sec:overfitanddoubledescent} the curious phenomenon of ``epoch wise double descent". Here, models trained over data with label noise show \emph{two distinct periods of descent in test error} (i.e., improvement in generalization), the second of which occurring \emph{after} the model has memorized the noisy labels. Our empirical analysis shows that the second period of descent is caused by learning \emph{new patterns} in the data, rather than relearning old ones forgotten at the overfit stage. This phenomenon, we show empirically, is caused by the \emph{simultaneous learning} of general patterns learned from clean data (which enhances performance continuously as learning progresses), and irrelevant specific patterns learned from noisy data (which degrade performance). 

Based on these empirical observations, we propose in Section~\ref{sec:method} a method that can effectively reduce the forgetting of test data. The challenge is to reduce overfit when test accuracy decreases with time, and to further improve the test accuracy even when it does not decrease with time (i.e., when overfit does not occur in the traditional sense). Accordingly, we construct a new prediction method, which aims to combine knowledge gained both in mid-training and after training. More specifically, the method delivers a weighted average of the class probability output vector between the final model and a set of checkpoints of the model from mid-training, where the checkpoints and their weights are chosen in an iterative manner using a validation dataset and our forget metric. The method is outlined in Alg.~\ref{alg:test}, see App~\ref{app:valalg}.

In Section~\ref{sec:empirical} we describe the empirical validation of our method in a series of experiments over image classification of datasets with and without label noise, using various network architectures, including in particular modern networks over Imagenet, thus showing that our method is universally useful. When compared with alternative methods that use the network's training history, our method shows comparable or better performance, while being more general and easy to use (both in implementation and hyper-parameter tuning). Specifically, in contrast with other methods, it does not depend on additional training choices that require much more time and effort to tune the new hyper-parameters.

\paragraph{Our main contributions}  \begin{inparaenum}[(i)] \item Novel perspective on overfit, entailing new scores to assess local overfit. \item Empirical evidence that overfit occurs "locally" even without a decrease in overall generalization. \item A simple and effective method to reduce overfit.\end{inparaenum}

\section{Related Work}

\paragraph{Studies of overfit and double descent} Overfit in deep neural networks, and specifically the double descent and epoch-wise double descent phenomena, has garnered increasing attention in recent years \citep[see recent review by][]{ganaie2022ensemble}. Double descent with respect to model size has been studied empirically in  \citep{belkin2019reconciling, nakkiran2021deep}, while epoch-wise double descent (which is the phenomenon analyzed here) was studied in \citep{stephenson2021and, heckel2020early}. These studies analyzed when and how epoch-wise double descent occurs, specifically in data with label noise, and explored ways to avoid it (sometimes at the cost of reduced generalization). In contrast, our research delves into the concept of double descent by investigating the extent to which neural networks forget. Our findings reveal that such phenomena are present, to some degree, even in datasets without label noise. We then use our observation to improve performance at \emph{inference time}, rather than change the training scheme as done in most previous work. Our line of study is complementary to - and should not be confused with -  the study of "benign overfitting", e.g., the fact that models can achieve perfect fit to the train data while still obtaining good performance over the test data.

\paragraph{Study of forgetting in prior work} Most relevant studies focus on the forgetting of training data, namely, the fact that some training points are memorized early on, but are then forgotten. This may occur when the network is not able to memorize all the training set. \citep{toneva2018empirical}, for example, analyzed the forgetting of points in the \emph{train data}, which is then used to score the importance of individual datapoints for training. In contrast, our work focuses on the forgetting of test points, which cannot be verified during training since the label of these points is not known. Another phenomenon, which may be confused with the one we are discussing, is "catastrophic forgetting" \citep{mccloskey1989catastrophic, ratcliff1990connectionist}. This occurs in a \emph{continual learning} scenario when the training data changes with time and the network does not have access to training data encountered at the beginning of training. In the scenario considered here this problem does not exist, since data does not expire.
 
\paragraph{Ensemble learning} Ensemble learning has been studied in machine learning for decades \citep{polikar2012ensemble}, including many recent works that employ deep neural networks ensembles, see \citep{ganaie2022ensemble, yang2023survey} for recent surveys. As ensembles are expensive, many works attempt to reduce their cost, specifically their training cost \cite{ganaie2022ensemble}. This line of works, to which our work belongs, is called "implicit ensemble learning", in which only a single network is learned in a way that "mimics" ensemble learning. A notable work in this field is  dropout \citep{srivastava2014dropout} and its variants, where some parts of the network are dropped at random during training, creating multiple "independent" networks inside the network. 

Utilizing checkpoints from the training history as a 'cost-effective' ensemble has also been considered. This was achieved by either considering the last epochs and averaging their probability outputs \citep{xie2013horizontal}, or by employing exponential moving average (EMA) on all the weights throughout training \citep{polyak1992acceleration}. While the latter method has demonstrated some success in reducing overfit, it has been reported to fail in some cases \citep{izmailov2018averaging}.


A number of methods \citep{izmailov2018averaging, garipov2018loss, huang2017snapshot} adopted a somewhat different approach that involves changing the training protocol of the network, such that it will converge to several local minima throughout training. These multiple solutions are then combined to form an ensemble classifier.  
These methods show great promise, impacting a range of fields such as medical applications \citep{nguyen2020lung,annavarapu2021deep}, fault diagnosis \citep{wen2019new}, attack detection \citep{rouzbahani2021snapshot} and land use classification \citep{noppitak2022dropcyclic}. However, these methods should be used with care, as they require the tuning of two inter-connected training schemes (the old and the new), and sometimes prolong the training time substantially at a potentially large financial cost in practical settings. Additionally, the new training scheme is not guaranteed to produce good and diverse local minima, and can even hurt performance as reported by \citet{guo2023stochastic}.  We conducted comprehensive comparisons to these methods (see Table~\ref{table:specialmethods}), demonstrating that in all instances, our approach either matches or outperforms them, all the while maintaining a significantly simpler design.

\begin{figure*}[h]
    \centering
    \begin{subfigure}[b]{0.24\linewidth}
        \includegraphics[width=\linewidth]{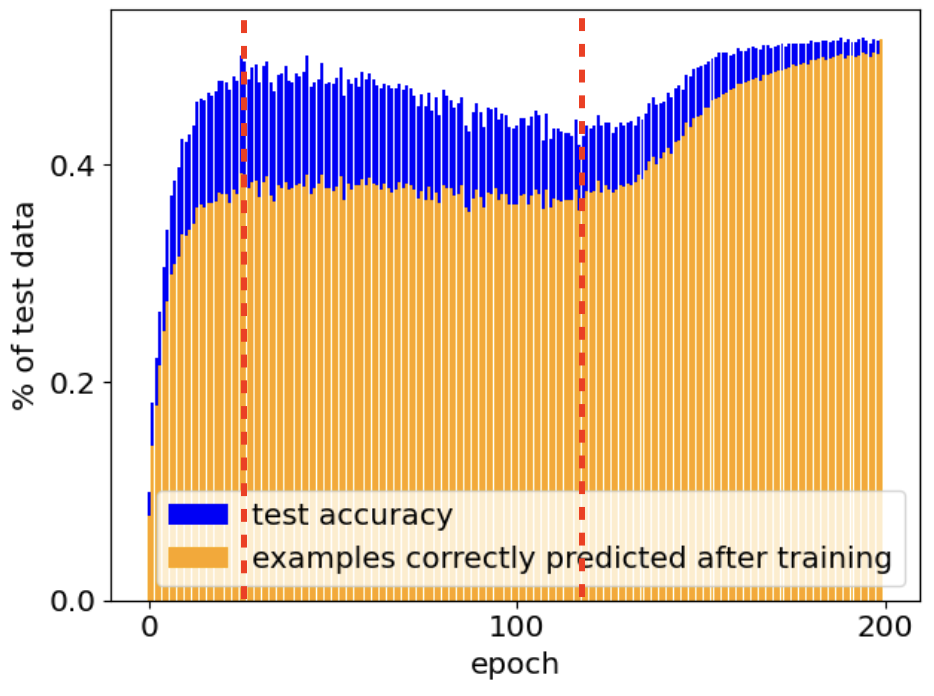}
    \vspace{-.6cm}
        \caption{Cifar100, 20\% sym noise}
        \label{subfig:DDc100sym20}
    \end{subfigure}
    \begin{subfigure}[b]{0.24\linewidth}
        \includegraphics[width=\linewidth]{figures/newtestbarTimgsym20r18.png}
    \vspace{-.6cm}
        \caption{TinyImagenet, 20\% sym noise}
        \label{subfig:DDTimgsym20}
    \end{subfigure}
    \begin{subfigure}[b]{0.24\linewidth}
        \includegraphics[width=\linewidth]{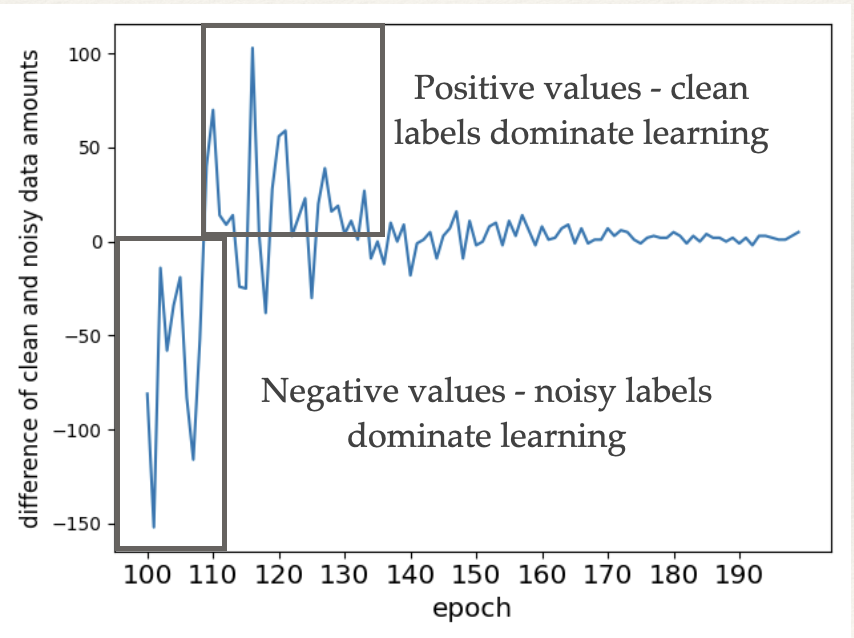}
    \vspace{-.6cm}
        \caption{Cifar100, 20\% sym noise}
        \label{subfig:c100sym20testacc}
    \end{subfigure}
    \begin{subfigure}[b]{0.24\linewidth}
        \includegraphics[width=\linewidth]{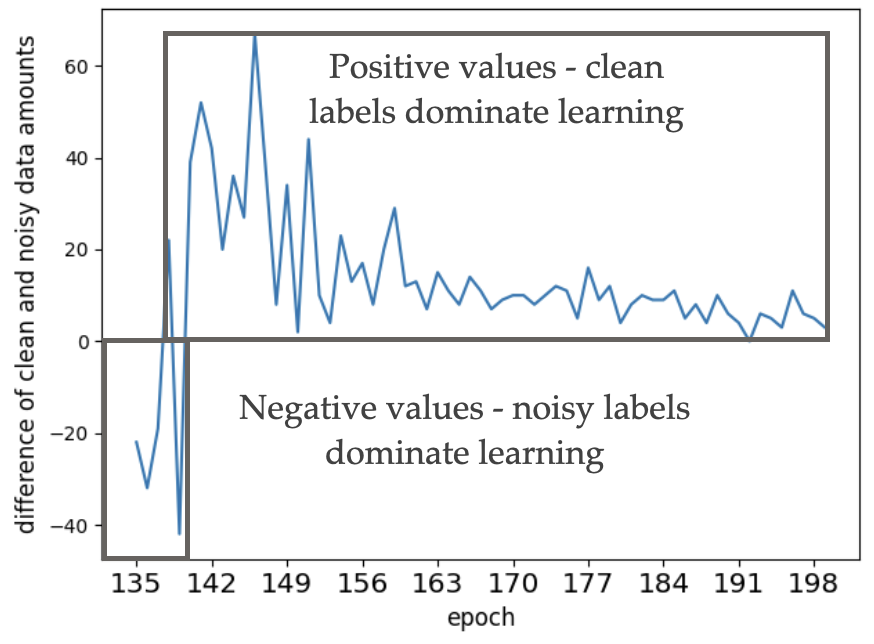}
    \vspace{-.6cm}
        \caption{TinyImagenet, 20\% sym noise}
        \label{subfig:logitsheatmapc100}
    \end{subfigure}

     \caption[heatmap]{(a)-(b): Blue denotes test accuracy. Among those correctly recognized in each epoch $e$, orange denotes the fraction that remains correctly recognized at the end. The test accuracy (the blue curve) shows a clear double ascent of accuracy, which is much less pronounced in the orange curve. During the decrease in test accuracy - the range of epochs between the first and second dashed red vertical lines - the large gap between the blue and orange plots indicates the fraction of test data that has been correctly learned in the first ascent and then forgotten, without ever being re-learned in the later recovery period of the second ascent. (c)-(d): Comparison between the amounts of the clean and noisy data with large loss (above a fixed threshold) at each epoch just before and during the second ascent of test accuracy (the range of epochs after the second dashed red vertical line). A positive value indicates that there are more clean data examples with large loss at this epoch, indicating that the model will now learn more general and correct patterns than wrong pattern caused by the noisy labels.}   
     \label{fig:doubledescent}
     \vspace{-.3cm}
\end{figure*}

\section{Overfit Revisited}
\label{sec:overfitanddoubledescent}


The textbook definition of overfit entails the co-occurrence of elevated train accuracy and reduced generalization. Let $acc(e,S)$ denote the accuracy over set $S$ in epoch $e$ - some epoch in mid-training, $E$ the total number of epochs, and $T$ the test dataset. Using test accuracy to approximate generalization, this implies that overfit occurs at epoch $e$ when $acc(e,T) \geq acc(E,T)$. However, test accuracy is a global measure that may obscure more subtle expressions of overfit, for example, when overfit only occurs in sub-regions of a continually evolving feature space. In this section, we attempt to expand our view of overfit, and develop a more sensitive metric that can capture the second form of overfit even in the absence of the first one. 

We begin with the observation that portions of the test data $T$ may be 'forgotten' by the network during training. We argue that this phenomenon indicates a form of local overfit, which can persist and negatively impact specific sub-regions of the dataset even as overall test accuracy continues to improve. Based on this observation, we propose to expand the  definition of overfit. This broader definition gains further validation from our examination of the 'epoch-wise double descent' phenomenon, which frequently occurs during training on datasets that contain significant label noise in the training set. In such cases, a notable forgetting of the test data coincides with the memorization of noisy labels, serving as an objective indicator of overfit. The primary implication of this analysis leads to a surprising revelation: even when training modern networks on standard datasets (devoid of label noise), where overfit (as traditionally defined) does not manifest,  \emph{the networks still appear to forget certain sub-regions of the test population}. This observation, we assert, signifies a significant and more subtle form of overfit.

\paragraph{Looking at overfit in a new way} Let $M_{e}$ denote the subset of test data mislabeled by the network at some epoch e. We now define two scores - \emph{learning} $L_e$ and \emph{forgetting} $F_e$ - as follows:
\begin{equation}
\label{eq:forget}
    F_e = acc(e,M_{E})\times\frac{|M_{E}|}{|T|}, \quad
    L_e = acc(E,M_{e})\times\frac{|M_{e}|}{|T|}
\end{equation}
$F_e$, referred to henceforth as the \emph{forget fraction}, represents the fraction of the test data that is correctly classified at epoch $e$ but incorrectly classified at the end of training (epoch $E$). This subset of the test data is known at epoch $e$, but later forgotten. On the other hand, $L_e$ denotes the fraction that will be known post-training but is misclassified at epoch $e$. Clearly $acc(E,T) = acc(e,T) + L_{e} - F_{e}$. Therefore, in line with the classical definition of overfit, if $L_{e} < F_{e}$, overfit indeed occurs sometime after epoch $e$.

\paragraph{What happens in the absence of classical overfit?} If $L_e \geq F_e~\forall e$, then by its classical definition \emph{overfit does not occur} since the test accuracy doesn't ever decrease. Nevertheless, there might still be numerous test examples that the final model misclassifies, even though they were classified correctly at some intermediate stage of training. This happens if at some epoch $e$ $L_e > F_e$ is still true, but $F_e$ is nevertheless large. As we show later on, this phenomenon is frequent among neural networks, more so than the traditionally defined overfit, which makes our definition useful in capturing this type of ill behavior.

\paragraph{Reflections on the epoch-wise double descent phenomenon} Epoch-wise double descent (see Fig.~\ref{fig:doubledescent}) is an empirical observation \citep{belkin2019reconciling}, which shows that neural networks can improve their performance even after overfitting, thus causing \emph{double descent in test error} during training (note that we show the corresponding \emph{double-ascent in test accuracy}). This phenomenon is characteristic of learning from data with label noise, and is strongly related to overfit since the dip in test accuracy co-occurs with the memorization of noisy labels. 

We examine the behavior of the novel score $F_e$ in this context and make a novel revelation: when we focus on the fraction of data acquired by the network during the second rise in test accuracy, we observe that the data newly memorized during these epochs often differs from the data forgotten during the overfit phase (the dip in accuracy). In fact, most of this data has been previously misclassified (refer to Figs.\ref{subfig:DDc100sym20} and \ref{subfig:DDTimgsym20}). To bolster this observation, Figs.\ref{subfig:c100sym20testacc} and \ref{subfig:logitsheatmapc100} further illustrate that during the later stages of training on data with label noise, when the second increase in test accuracy occurs, the majority of the data being memorized is, in fact, data with clean labels. 


\paragraph{What happens when there is no label noise?} When training deep networks on visual benchmark datasets without added label noise, double descent  rarely occurs, if ever. 

\begin{figure}[h]
    \centering
    \begin{subfigure}[b]{0.45\linewidth}
        \includegraphics[width=\linewidth]{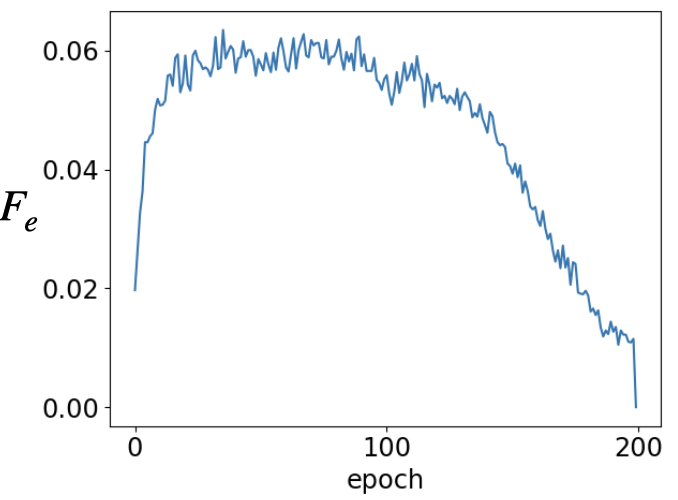}
    \vspace{-.5cm}
        \caption{TinyImagenet}
        \label{subfig:r18forgetTimg}
    \end{subfigure}
    \begin{subfigure}[b]{0.45\linewidth}
        \includegraphics[width=\linewidth]{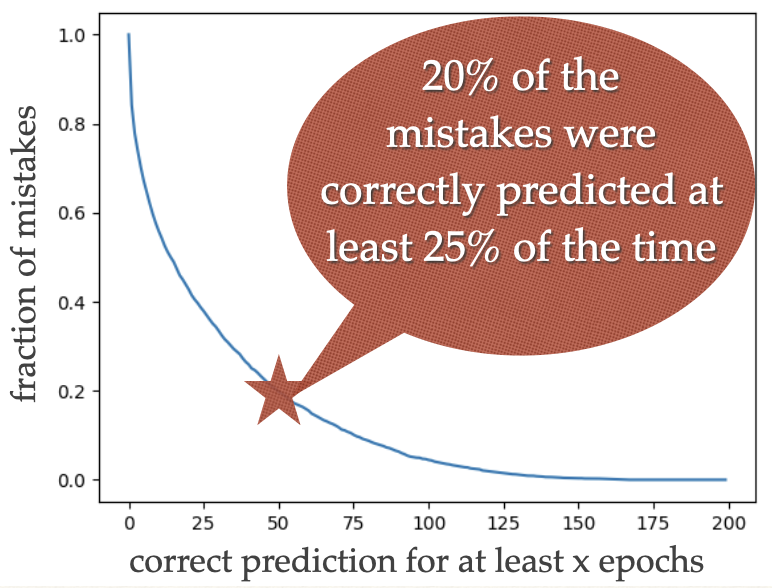}
    \vspace{-.5cm}        
    \caption{TinyImagenet}
        \label{subfig:cdfforgetr18Timg}
    \end{subfigure}
    \begin{subfigure}[b]{0.45\linewidth}
        \includegraphics[width=\linewidth]{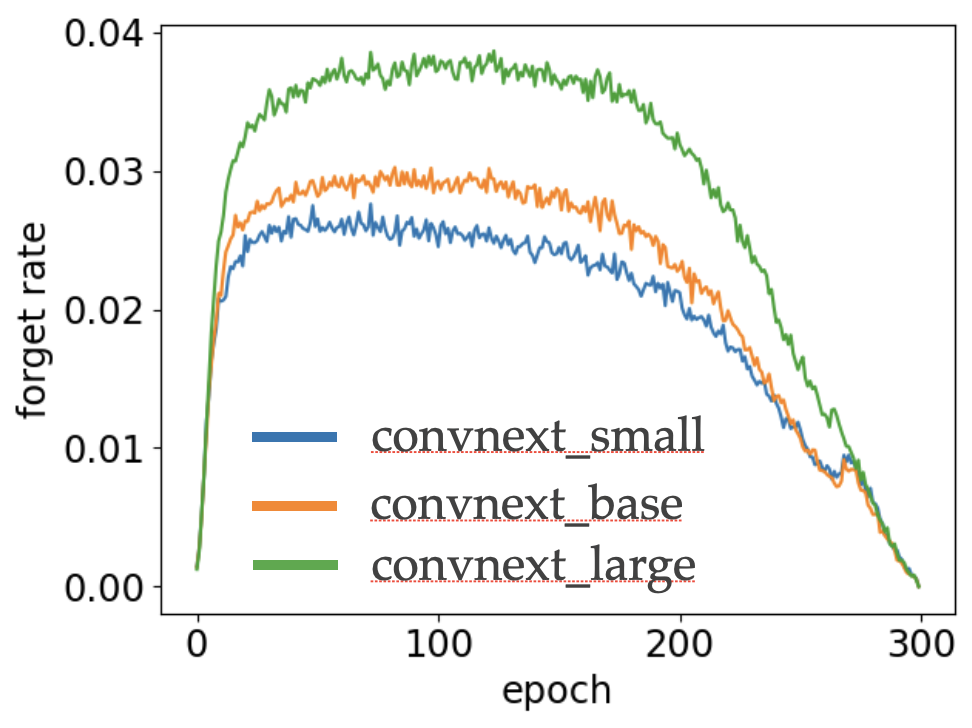}
    \vspace{-.5cm}
        \caption{ImageNet}
        \label{subfig:modelsizeforget}
    \end{subfigure}
    \begin{subfigure}[b]{0.45\linewidth}
        \includegraphics[width=\linewidth]{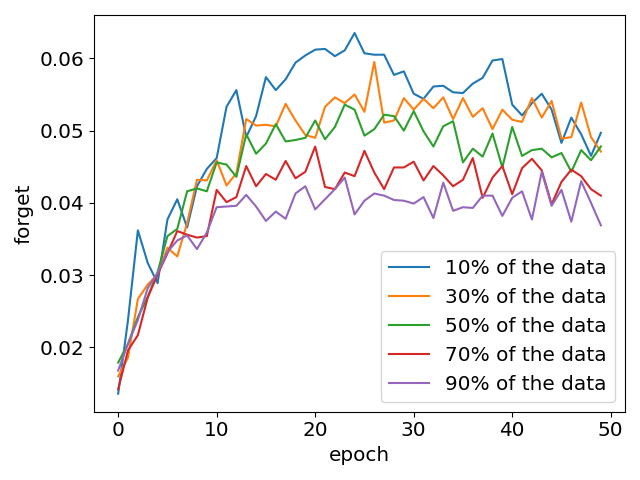}
    \vspace{-.5cm}
        \caption{Cifar100}
        \label{subfig:trainsizeforget}
    \end{subfigure}
     \caption[forget]{(a) The forget fraction $F_e$, as defined in (\ref{eq:forget}), of Resnet18 trained on TinyImagenet. (b) Within the set of wrongly classified test points after training, we show the fraction that was correctly predicted (y-axis) for at least x epochs (x-axis). (c) Comparison of the $F_e$ score of ConvNeXt trained on Imagenet, evaluated in three network sizes: small $\rightarrow$ blue, base $\rightarrow$ orange and large $\rightarrow$ green. Clearly, $F_e$ increases with the size of the network. (d) The $F_e$ score of Resnet18 trained on $10/30/50/70/90\%$ of the train data in cifar100 (purple/red/green/yellow/blue line, respectively) in the first 50 epochs of training (after which the score decreases); $F_e$ is significantly larger for the smallest set of only $10\%$.} 
     \label{fig:forgetrate}
\end{figure}

What about our new score $F_e$? To answer this question we trained various neural networks (ConvNets: Resnet, ConvNeXt; Visual transformers: ViT, MaxViT) on various datasets (CIFAR100, TinyImagenet, Imagenet) using a variety of optimizers (SGD, AdamW) and learning rate schedulers (cosine annealing, steplr). In Fig.~\ref{subfig:r18forgetTimg} and App~\ref{app:moreforget} we report the results, showing that all networks forget some portion of the data during training as in the label noise scenario, even if the test accuracy never decreases. Fig.~\ref{subfig:cdfforgetr18Timg} shows that this effect is not arbitrary: many examples have been correctly classified for a large portion of the network's training. 

In Figs.~\ref{subfig:modelsizeforget} and \ref{subfig:trainsizeforget} we connect our results to overfit. We show that when investigating either larger models or relatively small amounts of train data, which are scenarios that are expected to increase overfit based on theoretical considerations, both are associated with larger \emph{forget fraction} $F_e$. 

To further validate our results, we examined in App~\ref{app:moreforget} a different definition of "forget", in which we look at the last epoch in which an example was classified correctly. Interestingly, we see a dense span of epochs in which many examples are being classified correctly for the last time, which hints that subregions of the data population were "forgotten" in those epochs. In the appendix we also analyze the number of times a mistake of the last epoch was previously classified correctly, and show that for many examples, the correct classification was not a random effect but rather a consistent trend in significant parts of training.

\textbf{In summary,} our observations add up to the following: Neural networks can, and often will, "forget" significant portions of the test population as their training proceeds. In a sense, the networks \emph{are} overfitting, but this only occurs at some limited subregions of the world. The reason that overfit is not seen using the classical defintion, it appears, is that at the same time the networks still learn general patterns about the data from some correct - but hard to learn - train data, which allows for the test accuracy to keep improving. In the next section we discuss \textbf{how we can harness this observation to improve the network's performance}. 


\section{Method: Recovering Forgotten Knowledge}
\label{sec:method}

In Section~\ref{sec:overfitanddoubledescent} we showed that neural networks often show better performance in mid-training on a subset of the test data, even when the test accuracy is monotonically increasing with training epochs. Here we aim to integrate the knowledge obtained in mid and post training epochs, during inference time, in order to improve performance. To this end we must determine: \begin{inparaenum}[(i)] \item  which versions of the model to use; \item how to combine them with the post-training model; and \item how much weight to assign the post training model, whose overall performance is usually better. \end{inparaenum}

\begin{table*}[thb!]
\footnotesize
  \centering
  \begin{tabular}{l| c||c||c|c|c|c}
    \multicolumn{1}{ l |}{Method/\textbf{Dataset}} & \multicolumn{1}{ c ||}{\textbf{CIFAR-100}} & \multicolumn{1}{ c || }{\textbf{TinyImagenet}} & \multicolumn{4}{ c }{\textbf{Imagenet}} 
    \\ 
    \multicolumn{1}{ r |}{architecture} &     Resnet18 & Resnet18 & Resnet50  &  ConvNeXt large & ViT16 base & MaxViT tiny \\
    \toprule
        \emph{single network}   & 
        $78.07 \pm .28$ & $64.95 \pm .24$ & $75.74 \pm .14$ & $82.92 \pm .11$ & $79.16 \pm .1$ & $82.51 \pm .15 $ \\
        \hline
        \emph{horizontal (i)}   & $78.15 \pm .17$ & $64.89 \pm .18$ & $\mathbf{76.46 \pm .14}$ & $83.13 \pm .1$ & $79.11 \pm .1$ & $82.77 \pm .1$ \\
        \emph{fixed jumps (i)}   & $78.04 \pm .23$ & $ 66.54 \pm .35$ & $75.5 \pm .09$ & $82.37 \pm .1$ & $78.67 \pm .08$ & $83.38 \pm .1$ \\
        \emph{KF (ours) (i)}   & $\mathbf{78.33 \pm .08}$ & $\mathbf{66.98 \pm .37}$ & $75.88 \pm .14$ & $83.18 \pm .16$ & $\mathbf{79.93 \pm .11}$ & $83.34 \pm .04$ \\

        \hline
        \emph{horizontal $(\infty)$}   & $78.23 \pm .17$ & $65.11 \pm .3$ & $\mathbf{76.42 \pm .1}$ & $83.02 \pm .06$ & $79.53 \pm .13$& $82.93 \pm .14$  \\
        \emph{fixed jumps $(\infty)$}   & $\mathbf{79.17 \pm .08}$ & $68.24 \pm .38$ & $75.72 \pm .18$ & $\mathbf{83.86 \pm .06}$ & $79.11 \pm .13$ & $\mathbf{83.78 \pm .15}$ \\

        \emph{KF (ours) $(\infty)$} & $\mathbf{79.13 \pm .14}$ &$\mathbf{68.5 \pm .36}$ & $\mathbf{76.52 \pm .16}$ & $\mathbf{83.96 \pm .09}$ & $\mathbf{80.34 \pm .08}$ & $\mathbf{83.81 \pm .14}$\\
        \bottomrule
        \emph{improvement} & $\mathbf{1.05 \pm .14}$ &$\mathbf{3.54 \pm .14}$ & $\mathbf{.78 \pm .04}$ & $\mathbf{1.03 \pm 13}$ & $\mathbf{1.17 \pm .08}$ & $\mathbf{1.29 \pm .02}$\\      
  \end{tabular}
  \vspace{0.2cm}
  \caption{Mean (over random validation/test splits) test accuracy (in percent) and standard error on image classification datasets, comparing our method and  baselines described in the text. The last row shows the improvement of the best performer over the single network. Suffixes: $(i)$ denotes a limited budget scenario, in which we use our method in a non-iterative manner; $(\infty)$ denotes the unlimited budget scenario, where we use our iterative version of the method. In each case, the baselines employ the same  number of checkpoints as our method. } 
  \label{table:regularnetworks}
\end{table*}

\begin{table*}[thb!]
\footnotesize
  \centering
  \begin{tabular}{l| c|| c|c|c || c|c || c|c}
    \multicolumn{1}{ c |}{Method/\textbf{Dataset}} & \multicolumn{1}{ c ||}{\textbf{Animal10N}} & \multicolumn{3}{ c ||}{\textbf{CIFAR-100 asym}} & \multicolumn{2}{ c || }{\textbf{CIFAR-100 sym}} & \multicolumn{2}{ c }{\textbf{TinyImagenet}} 
    \\ 
    \multicolumn{1}{ r |}{\% label noise} &    8\% & 10\% & 20\% & 40\%  &  20\% & 40\% &  20\% & 40\%\\
    \toprule
    \emph{single network}   &  $85.9\pm.3$ & $72.1 \pm .1$ & $67.1 \pm .5$ & $49.4 \pm .3$& $65.4 \pm .3$ & $56.9 \pm .1$ &  $56.2 \pm .2$ & $49.8 \pm .3$ \\

        \hline
    \emph{fixed jumps $(\infty)$}   & $87.1\pm.4$ & $76.2 \pm .1$ & $73.9 \pm .1$ & $59.9 \pm .6$& $72.8 \pm .1$ & $66.5 \pm .1$ & $60.0 \pm .8$ & $54.16 \pm .3$ \\
    \emph{horizontal $(\infty)$}  & $86.3\pm.3$ & $75.4 \pm .3$ & $73.4 \pm .1$ & $58.5 \pm .1$& $71.1 \pm .38$ & $65.2 \pm .1$ & $59.3 \pm .3$ & $51.7  \pm .2$ \\
    \emph{KF (ours) $(\infty)$} & $\mathbf{87.8\pm.4}$ & $\mathbf{76.6 \pm .3}$ & $\mathbf{74.2 \pm .1}$& $\mathbf{62.1 \pm .5}$& $\mathbf{72.8 \pm .1}$ & $\mathbf{67.0 \pm .1}$ & $\mathbf{62.8 \pm .2}$ &$\mathbf{57.0 \pm .5}$ \\
    \bottomrule
    \emph{improvement} & $ \mathbf{1.9\pm.4}$ & $\mathbf{4.4 \pm .2}$ & $\mathbf{7.1 \pm .6}$& $\mathbf{12.6 \pm .2}$& $\mathbf{7.4 \pm .4}$ & $\mathbf{10.1 \pm .1}$ & $\mathbf{6.6 \pm .1}$ &$\mathbf{7.2 \pm .1}$ \\

  \end{tabular}
  \vspace{0.2cm}
  \caption{Mean test accuracy (in percent) and standard error of Resnet 18, comparing our method and the baselines on datasets with large label noise and significant overfit. We include a comparison to the Animal10N dataset, which has innate label noise. Note that as is customary, only the train data has label noise while the test data remains clean for a fair evaluation.} 

  \label{table:labelnoise}
\end{table*}

\begin{table*}[thb!]
\footnotesize
  \centering
  \begin{tabular}{l| c || c || c || c|c || c|c}
    \multicolumn{1}{ c |}{Method/\textbf{Dataset}} & \multicolumn{1}{ c ||}{\textbf{CIFAR-100}}& \multicolumn{1}{ c ||}{\textbf{\tiny TinyImagenet}}&\multicolumn{1}{ c ||}{\textbf{\tiny Animal10N}}& \multicolumn{2}{ c ||}{\textbf{CIFAR-100 asym}} & \multicolumn{2}{ c  }{\textbf{CIFAR-100 sym}} 
    \\ 
    \multicolumn{1}{ r |}{\% label noise} &     0\% & 0\% & 8\% & 20\% & 40\%  &  20\% & 40\% \\
    \toprule
    \emph{FGE $(\infty)$}   & $78.9 \pm .4$& $67.7\pm.1$ & $86.5\pm0.6$ &$67.1 \pm .2$& $48.1 \pm .3$& $66.5 \pm .1$& $52.1 \pm .1$\\
    \emph{SWA $(\infty)$}   & $78.8 \pm .1$& $\mathbf{69.3 \pm .6}$ & $\mathbf{88.1\pm.2}$& $66.6 \pm .1$& $46.9 \pm .2$& $65.6 \pm .4$& $50.0 \pm .1$\\
    \emph{snapshot $(\infty)$}   & $78.4 \pm .1$ & $\mathbf{69.3 \pm .4}$& $86.8\pm.3$ &  $72.1 \pm .4$ & $52.8 \pm .6$& $70.8 \pm .5$ & $63.8 \pm .2$  \\
    \emph{KF (ours) $(\infty)$} & $\mathbf{79.3 \pm .2}$ & $\mathbf{69.4 \pm .6}$& $\mathbf{87.8\pm.4}$& $\mathbf{74.2 \pm .1}$& $\mathbf{62.1 \pm .5}$& $\mathbf{72.8 \pm .1}$ & $\mathbf{67.0 \pm .1}$\\
  \end{tabular}
  \caption{Mean test accuracy (in percent) and standard error of Resnet18, comparing our method and baseline methods that alter the training.} 
\vspace{-0.3cm}
  \label{table:specialmethods}
\end{table*}

\paragraph{Choosing an early epoch of the network} Given a set of epochs $\{1,\ldots,E\}$ and corresponding forget rates $\{F_e\}_e$, we first single out the model $n_A$ obtained at epoch $A = argmax_{e \in \{1,...,E\}}F_e$. This epoch is most likely to correct mistakes of the model on "forgotten" test data.



\paragraph{Combining the predictors} How to combine the two models, $n_A$ where the forget fraction is maximal, and $n_E$ where train accuracy is maximal? A common practice is to average models' output. However, since the performance of $n_E$ is typically better than that of $n_A$, we use a weighted average instead, giving $n_E$ a larger weight. This guarantees that our method will not harm the general performance, as it can always give a zero weight to the early checkpoint $n_A$.

\paragraph{Improving robustness} To improve our method's robustness to the choice of epoch $A$, we use a span of epochs around $A$, denoted by $\{n_{A-w},...,n_A,..., n_{A+w}\}$. The vectors of probabilities computed by each checkpoint are averaged before forming an ensemble with $n_E$. In our experiments, we use a fixed width $w=1$.

\paragraph{Working in an iterative manner} As we now have a new predictor, we can now find another alternative predictor from the training history that maximizes accuracy on the data misclassified by the new predictor, and combine their knowledge as described. This can be done iteratively, until no further improvement is achieved. 

\paragraph{Choosing hyper-parameters} In order to compute $F_e$ (for the early epoch choice) and to find the best weights and epoch span, we use a validation set, which is a part of the labeled data not shown to the model during initial training. This is done \textbf{post training} as it has no influence over the training process, and thus \emph{doesn't incur additional costs over the training time}. We follow common practice, and show in Section~\ref{abl:val} that after finding the best hyper-parameters it is possible to retrain the model on the complete training set and validation set and use our method \emph{without hurting performance}, and while maintaining our superiority over alternative methods also trained on the full data. 

We call our method \textbf{K}nowledge\textbf{F}usion (KF), and evaluate it in Section~\ref{sec:empirical}. Its pseudocode can be found in App~\ref{app:valalg}.




\section{Empirical evaluation}
\label{sec:empirical}

We now demonstrate the superior performance of our method as compared to the original predictor, i.e. the network after training, as well as other baselines. We evaluate our method using various image classification datasets, neural networks architectures, and training schemes. The main results of our empirical evaluation are presented in Tables~\ref{table:regularnetworks}-\ref{table:specialmethods}, followed by an extensive ablation study (and additional comparisons) in Section~\ref{sec:ablation}.

\paragraph{Review of empirical results} In Table~\ref{table:regularnetworks} we report the results of our method using multiple architectures trained on cifar100, TinyImagenet and Imagenet, with different learning rate schedulers and optimizers. For comparison, we report the results of both the original predictor and some simple baselines. We continue with additional experiments on settings connected to overfit in Table~\ref{table:labelnoise} and App~\ref{app:additionaleval}, where we test our methods on these datasets with injected symmetric and asymmetric label noise (see App~\ref{app:implementationdetails}), as well as on real label noise dataset (Animal10N). Note that as customary, the label noise exists only in the train data while the test data remains clean for model evaluation. 

In Table~\ref{table:specialmethods} and App~\ref{app:additionaleval} we compare our method to additional methods that adjust the training protocol itself, using both clean and noisy datasets. We employ these methods using the same network architecture as our own, after suitable hyper-parameter search
. Finally, we compare in App~\ref{app:additionalablation} (Fig.~\ref{fig:oursvsensemble}) our method with an ensemble of independent networks, to evaluate how much of the ensemble's performance boost can be gained using our method (without the extra cost of ensemble training), see details in App~\ref{app:implementationdetails}.

In each experiment we use half of the \emph{test data} for validation, to compute our method's hyper-parameters (the list of alternative epochs and $\{\epsilon_i\}$), and then test the result on the remaining test data. The accuracy reported here is only on the remaining test data, averaged over three random splits of validation and test data, using different random seeds. In Section~\ref{abl:val} we show that when data is limited, we can train a network on a subset of the training data while using the left out data for hyper-parameter tuning.  As customary, these same parameters are later used with models trained on the full data, demonstratively without deteriorating the results.

\paragraph{Baselines} Our method incurs the training cost of a single model, and thus, following the methodology of \citep{huang2017snapshot}, we compare ourselves to methods that require the same amount of training time. Our baselines are from two groups of methods. The first group includes methods that do not alter the training process:
\begin{itemize}[leftmargin=0.65cm]
    \item \textbf{Single network}: the original network, after training. 
    \item \textbf{Horizontal ensemble} \citep{xie2013horizontal}: this method uses a set of epochs at the end of the training, and delivers their average probability outputs (with the same number of checkpoints as we do). 
    \item \textbf{Fixed jumps}: this baseline was used in \citep{huang2017snapshot}, where several checkpoints of the network, equally spaced through time, are taken as an ensemble.
\end{itemize} 
The second group includes methods that \emph{alter} the training protocol. While this is not a directly comparable set of methods, as they focus on a complementary way to improve performance, we report their results in order to further validate the usefulness of our method. This group includes the following methods:
\begin{itemize}[leftmargin=0.65cm]
    \item \textbf{Snapshot ensemble} \citep{huang2017snapshot}: in this method the network is trained in several "cycles", each ending with a large increase of the learning rate that pushes the network away from the local minimum. The network is meant to converge to several different local minima during training, which are used as an ensemble. 
    \item \textbf{Stochastic Weight Averaging} (SWA) \citep{izmailov2018averaging}: in this method the network is regularly trained for a fixed training budget of epochs, and is then trained using a circular/constant learning rate to converge to several local minima, whose weights are averaged to get the final predictor. To achieve fair comparison in training budget, we train the network using our training method for 75\% of the epochs, followed by their unique training for the remaining 25\% epochs. 
    \item \textbf{Fast Geometric Ensembling} (FGE) \citep{garipov2018loss}: similar to SWA, except that the final predictor is constructed by averaging the probability outputs of each model, instead of their weights. In this comparison we match budgets as explained above.
\end{itemize} 

Comparisons to additional baselines that are relevant to resisting overfit, including early stopping and test time augmentation, are discussed in App~\ref{abl:ema}.

\section{Ablation Study}
\label{sec:ablation}



In this section we investigate some limitations and practical aspects of our method. In Section~\ref{abl:val} we show that a separate validation set is not really necessary for the method to work well. In Section~\ref{subsec:checkpointsnum} we investigate how many checkpoints are needed for the method to be effective, showing that only $5-10\%$ of the past checkpoints are sufficient. In Section~\ref{abl:subopttrain} we investigate the added value of our method when using only partial hyper-parameter search, as is common in real applications, which leads to sub-optimal training. Interestingly, our method is shown to be even more beneficial in the sub-optimal scenario, and reduces the gap between the optimal and sub-optimal networks. Finally, in Section~\ref{abl:transferlearning} we show our method is effective in transfer learning scenerio, when the network's initial weights are pretrained.


Additional evaluations are described in App~\ref{app:additionalablation}, where we show that: (i) our method is superior compared to exponantial-moving-average (EMA), early stopping and test time augmentation (App~\ref{abl:ema}); (ii) our method's improvement can grow as the number of parameters grow (App~\ref{subsec:modelsize}); (iii) a large portion the improvement of a regular ensemble of independent networks can often be obtained using our method at a much lower cost (App~\ref{subsec:regens}); and (iv) our method does not have negative effects on the model's fairness (App~\ref{abl:fairness}).

\subsection{Removing the requirement for validation set}
\label{abl:val}

In this experiment, we follow a common practice with respect to the validation data: we train our model on cifar100 and TinyImagenet using only 90\% of the train data, use the remaining 10\% for validation, and finally retrain the model on the full train data while keeping the same hyper-parameters for inference. The results are almost identical to those reported in Table~\ref{table:regularnetworks}. This validates the robustness of our method to the (lack of) a validation set. 

\subsection{number of checkpoints used}
\label{subsec:checkpointsnum}

\begin{figure}[h]
    \centering
    \begin{subfigure}[b]{0.45\linewidth}
        \includegraphics[width=\linewidth]{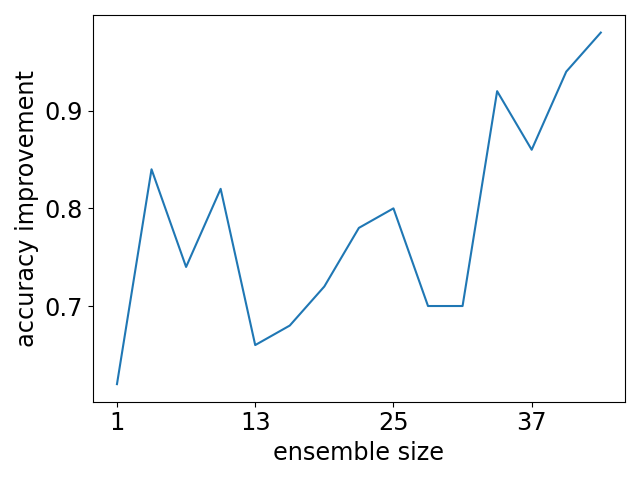}
    \vspace{-.5cm}
        \caption{cifar100}
    \end{subfigure}
    \begin{subfigure}[b]{0.45\linewidth}
        \includegraphics[width=\linewidth]{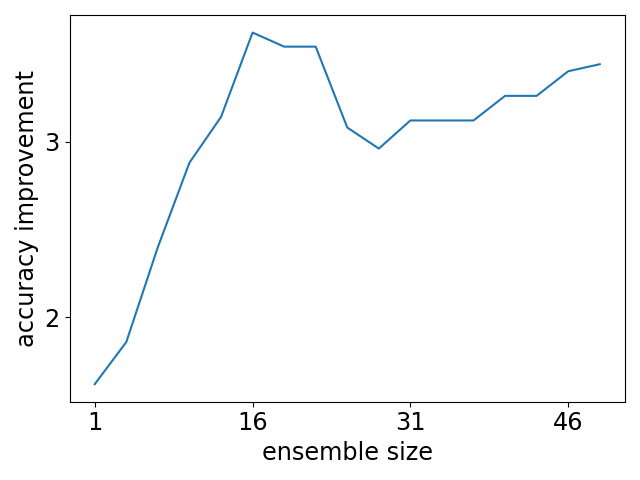}
    \vspace{-.5cm}
        \caption{TinyImagenet}
    \end{subfigure}
    \begin{subfigure}[b]{0.45\linewidth}
        \includegraphics[width=\linewidth]{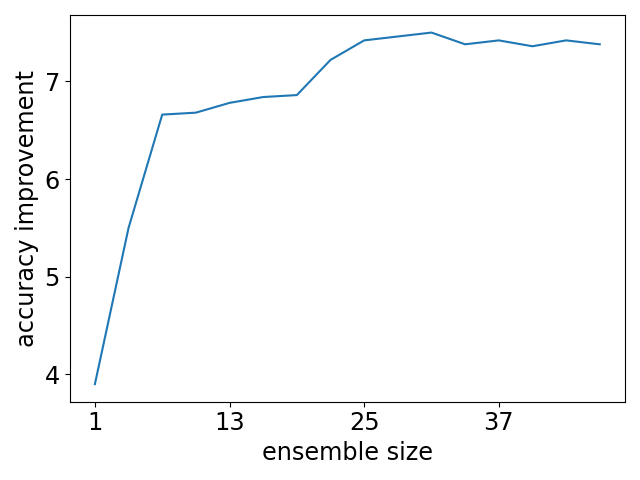}
    \vspace{-.5cm}
        \caption{cifar100, 20\% asym noise}
    \end{subfigure}
    \begin{subfigure}[b]{0.45\linewidth}
        \includegraphics[width=\linewidth]{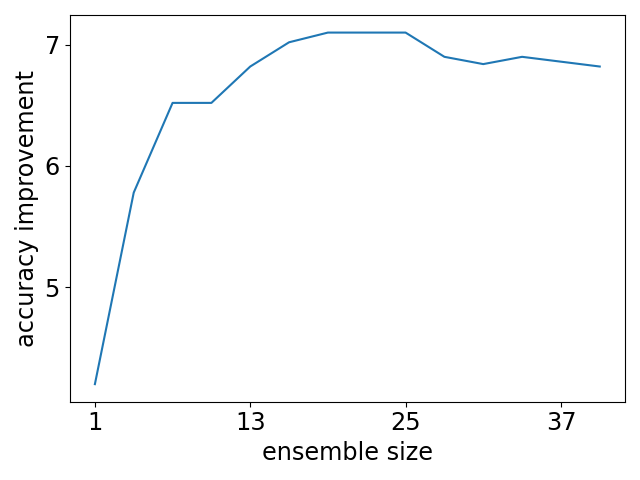}
    \vspace{-.5cm}
        \caption{TinyImagenet, 20\% sym noise}
    \end{subfigure}
     \caption[ensemble size]{Improvement achieved by our method when using a different number of checkpoints (shown on the x-axis).
     } 
     \label{fig:ensemblesize}
\end{figure}

Here we evaluate the cost entailed by the use of an ensemble at inference time. In Fig.~\ref{fig:ensemblesize} we report the improvement in test accuracy as compared to a single network, when varying the ensemble size. The results indicate that almost all of the improvement can be obtained using only $5-10\%$ of the checkpoints, making our method practical in real life.

\subsection{Optimal vs sub-optimal training}
\label{abl:subopttrain}

In real life, full search for the optimal training scheme and hyper-parameters is not always possible, leading to sub-optimal performance. Interestingly, our method can be used to reduce the gap between optimal and sub-optimal training, as seen in Table~\ref{table:suboptimal}, where the gap between optimally and suboptimally trained MaxViT over Imagenet reduces by almost half when applying our method (on both models). 

\begin{table}[h]
\footnotesize

        \centering
  \begin{tabular}{l| c||c}
    \multicolumn{1}{ c |}{training/\textbf{method}} & \multicolumn{1}{ c ||}{\textbf{original network}} & \multicolumn{1}{ c }{\textbf{KF}} 
    \\ 
    \toprule
    \emph{regular training}   & $82.5 \pm .1$ & $83.8 \pm .1$ \\
    \emph{sub-optimal training}   & $77.3 \pm .1$ & $81.0 \pm .1$\\
    \hline
    \emph{improvement}   & $\mathbf{5.2}$ &$\mathbf{2.8}$\\

  \end{tabular}
\caption{Mean test accuracy (in percent) and ste, over random validation/test split. MaxViT is trained to classify Imagenet, comparing optimal and sub-optimal training with and without KF.} 
    \vspace{-0.25cm}

\label{table:suboptimal}
  

\end{table}

\subsection{Transfer learning}
\label{abl:transferlearning}

Another popular method to improve performance and reduce overfit employs transfer learning, in which the model weights are initialized using pre-trained weights over a different task, for example Imagenet pretrained weights. This is followed by \emph{fine tuning} either the entire model or only some of its layers (its head for example). In Table~\ref{table:transferlearning} we show that our method is complementary to the use of transfer learning in both cases, as it can still improve performance in this scenario. Note that when finetuning only the last layers, our method is \textbf{almost free of overhead costs}, as one needs to save and use at inference most of the model only once. Thus, the only overhead involves the memory and inference costs of the different head checkpoints in finetuning, whose size is insignificant compared to the rest of the model.

\begin{table}[h]
\footnotesize

        \centering
  \begin{tabular}{l| c||c}
    \multicolumn{1}{ c |}{Method/\textbf{Dataset}} & \multicolumn{1}{ c ||}{\textbf{CIFAR-100}} & \multicolumn{1}{ c }{\textbf{TinyImagenet}} 
    \\ 
    \toprule
        \scriptsize{\emph{fully finetuned Resnet18} }  & $80.72 \pm .53$ & $75.01 \pm .12$ \\
        \scriptsize{\emph{fully finetuned Resnet18 + KF}}   & $\mathbf{81.64 \pm .27}$ & $\mathbf{75.6 \pm .18}$\\
        \scriptsize{\emph{partially finetuned Resnet18}}   & $61.7 \pm .60$ & $54.78 \pm .13$ \\
        \scriptsize{\emph{partially finetuned Resnet18 + KF} }  &  $\mathbf{65.24 \pm .67}$ & $\mathbf{59.5 \pm .03}$\\
  \end{tabular}
\caption{Mean test accuracy 
     over random validation/test split.  Our method applied to Resnet18 pre-trained on Imagenet, while finetuning the entire model (top) or only the head (bottom).} 

\label{table:transferlearning}

\end{table}

\section{Method: Discussion and Limitations}  

Our method provides a significant improvement of around $1\%$ on  modern neural networks over Imagenet, which typically implies more than $5\%$ reduction in test error. It is often complementary to other methods (such as EMA) that aim to reduce overfit, and often succeeds to reduce overfit when such methods fail (see Section~\ref{abl:transferlearning} and App~\ref{abl:ema}). Our method is especially useful in transfer learning settings when only a few layers are finetuned, as it: \begin{inparaenum}[(i)] \item  significantly improves performance; and \item \textbf{adds very little overhead in inference and memory costs}, since most of the model is saved used at inference time only once. \end{inparaenum}
    
In more difficult settings, such as a small network for complex data (e.g., Resnet18 over TinyImagenet) or datasets with label noise, our method improves performance even further, leading to a nice error reduction of ${\scriptstyle \sim} 15\%$ in the reasonable setting of $10\%$ asymmetric noise. This may be the case when dealing with complex and confusing natural datasets such as Animal10N  \citep{song2019selfie}.

When compared to baselines, we achieve comparable or better results (see, for example, our improvement in ViT16 over Imagenet and Resnet18 over TinyImagenet). Another large advantage of our method is that it is \textbf{independent of training choices}. In contrast, the horizontal method seems to show little improvement when the cosine-annealing learning rate scheduler is used (all experiments but Resnet50 over Imagenet), while fixed-jumps with no cycles \citep{huang2017snapshot} is known to show little improvement (or none at all) when step-size learning rate scheduler is used (Resnet50 over Imagenet in our experiments). Finally, while  the more complicated baselines proved more useful than the other baselines (but not more than our method), this was only made possible by an extensive hyper-parameter search.

While being simple and useful, our method has a few limitations as it requires: (i) validation data to tune the hyper-parameters; (ii) using multiple checkpoints, which incurs overhead in inference time and memory usage; (iii) the occurrence of forgetting to have any impact. These limitations can be mitigated, however, as shown in Section~\ref{sec:ablation}:
\begin{inparaenum}[(i)] \item A subset of the train set can be effectively used for validation. \item A few checkpoints can already achieve most of the method's benefit. We note that these checkpoints can run in parallel on multiple GPUs, and the memory overhead is not likely to be excessive for big datasets. For example, the Imagenet dataset weighs 167GB\footnote{https://www.kaggle.com/competitions/imagenet-object-localization-challenge/data.}, while a regular resnet50 weighs around 230MB. \item The method can revert to the original single network if no improvement on a validation set is seen, a benefit lacking in the methods listed in Table~\ref{table:specialmethods}, as it doesn't require any changes to the learning process. Lastly, our empirical evaluation shows that our method has no negative effects on the model's fairness, making it safe to use. \end{inparaenum}

\section{Conclusions and Future Work}

We revisited the problem of \emph{overfit} in deep learning, proposing to track the forgetting of validation data in order to detect local overfit. We connected our new perspective with the \emph{epoch wise double descent} phenomenon, empirically extending its scope while demonstrating that a similar effect occurs in benchmark datasets with clean labels. Inspired by these new empirical observations, we constructed a simple yet general method to improve classification at inference time. We then empirically demonstrated its effectiveness on many datasets and modern network architectures. The method improves modern networks by around 1\% accuracy over Imagenet, and is especially useful in some transfer learning settings where its benefit is large and its overhead is very small. In future work we will investigate and characterize the "forgotten" examples, and seek ways to achieve better and more effective combinations of  checkpoints.

\bigskip
\paragraph{Acknowledgement}
This work was supported by grants from the Israeli Council of Higher Education and the Gatsby Charitable Foundations.
\bigskip

\bibliographystyle{plain}
\bibliography{ArxivSubmission}

\appendix

\section*{Appendix}

\section{Additional forget fraction examples}
\label{app:moreforget}

\begin{figure*}[h]
    \centering
    \begin{subfigure}[b]{0.24\linewidth}
        \includegraphics[width=\linewidth]{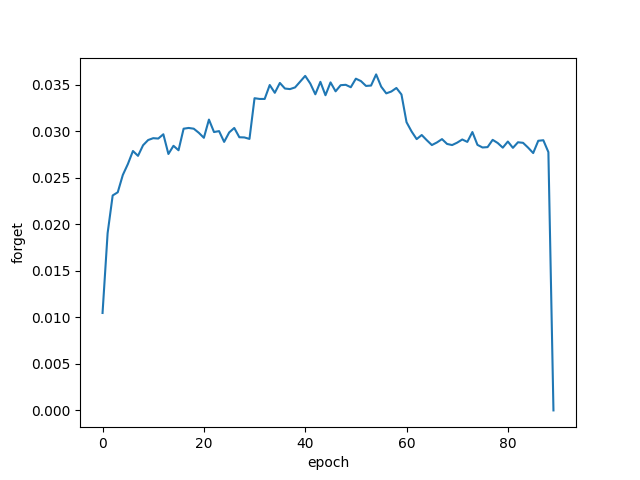}
    \vspace{-.5cm}
        \caption{Resnet50 on Imagenet,  steplr scheduler}
        \label{subfig:r18forgetImg}
    \end{subfigure}
    \begin{subfigure}[b]{0.24\linewidth}
        \includegraphics[width=\linewidth]{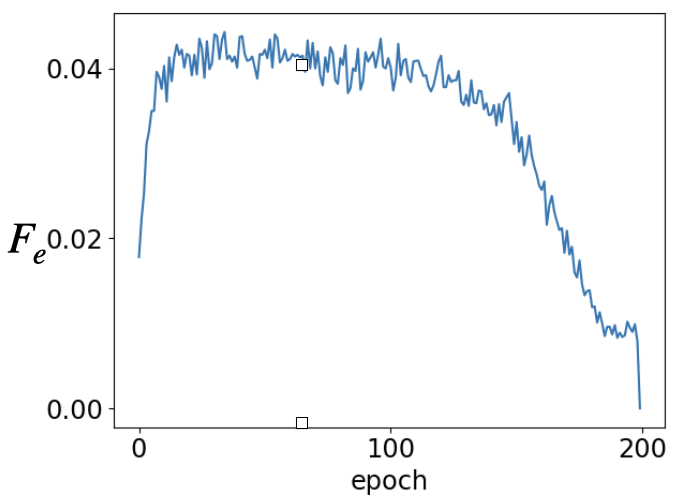}
    \vspace{-.5cm}        \caption{Densenet121 on cifar100}
        \label{subfig:ForgetDenseC100}
    \end{subfigure}
    \begin{subfigure}[b]{0.24\linewidth}
        \includegraphics[width=\linewidth]{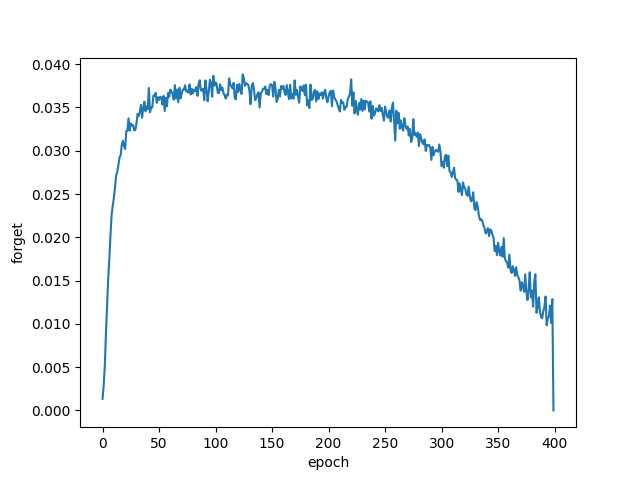}
    \vspace{-.5cm}
        \caption{MaxViT tiny on Imagenet}
        \label{subfig:MaxViTforgetImg}
    \end{subfigure}
    \begin{subfigure}[b]{0.24\linewidth}
        \includegraphics[width=\linewidth]{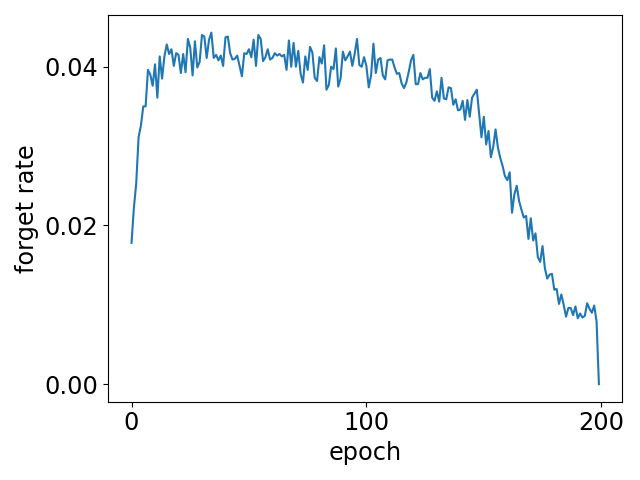}
    \vspace{-.5cm}
        \caption{Densenet121 on TinyImagenet}
        \label{subfig:Denseforgettiny}
    \end{subfigure}
     \caption[forget]{ The forget fraction $F_e$, as defined in (\ref{eq:forget}), of common neural networks trained on image classification datasets. } 
     \label{fig:moreforgetrate}
\end{figure*}

\begin{figure*}[h]
    \centering
    \begin{subfigure}[b]{0.24\linewidth}
        \includegraphics[width=\linewidth]{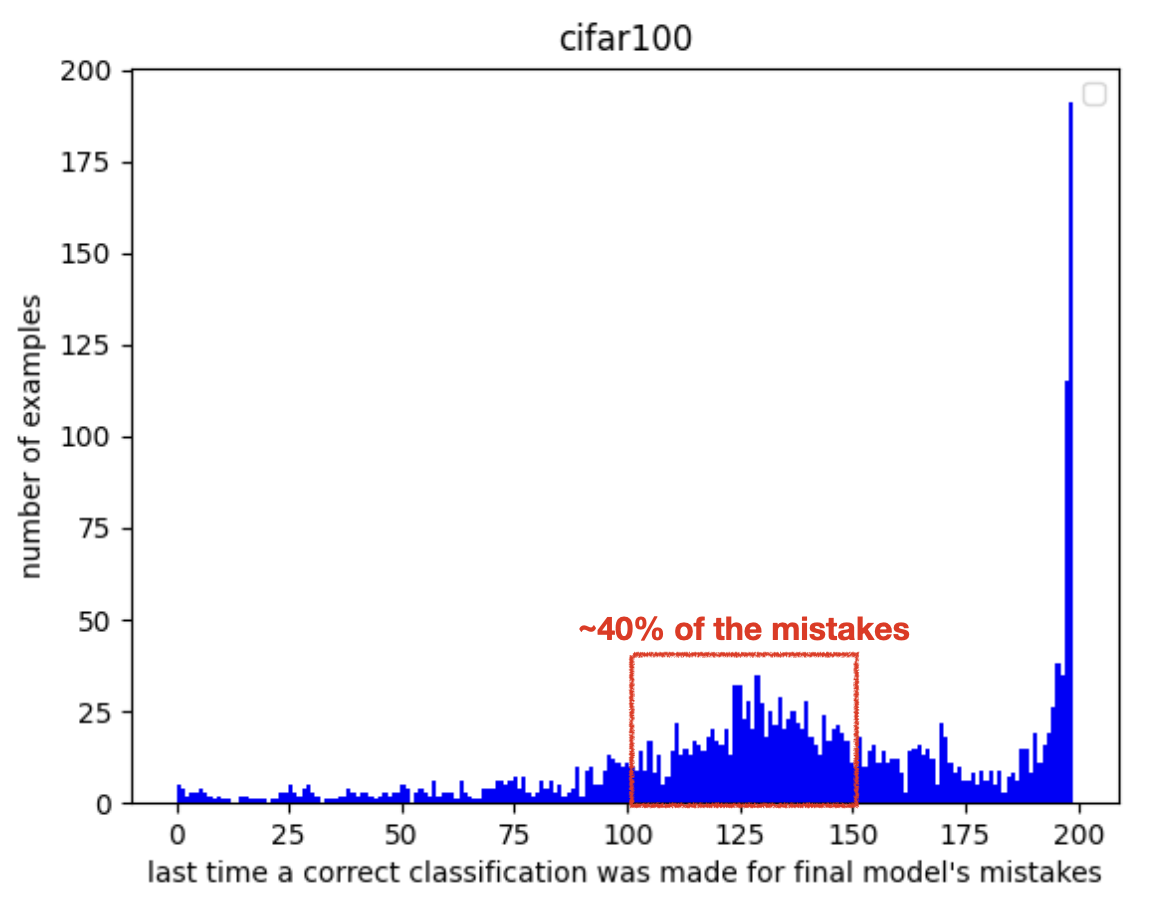}
    \vspace{-.5cm}
        \caption{cifar100}
    \end{subfigure}
    \begin{subfigure}[b]{0.24\linewidth}
        \includegraphics[width=\linewidth]{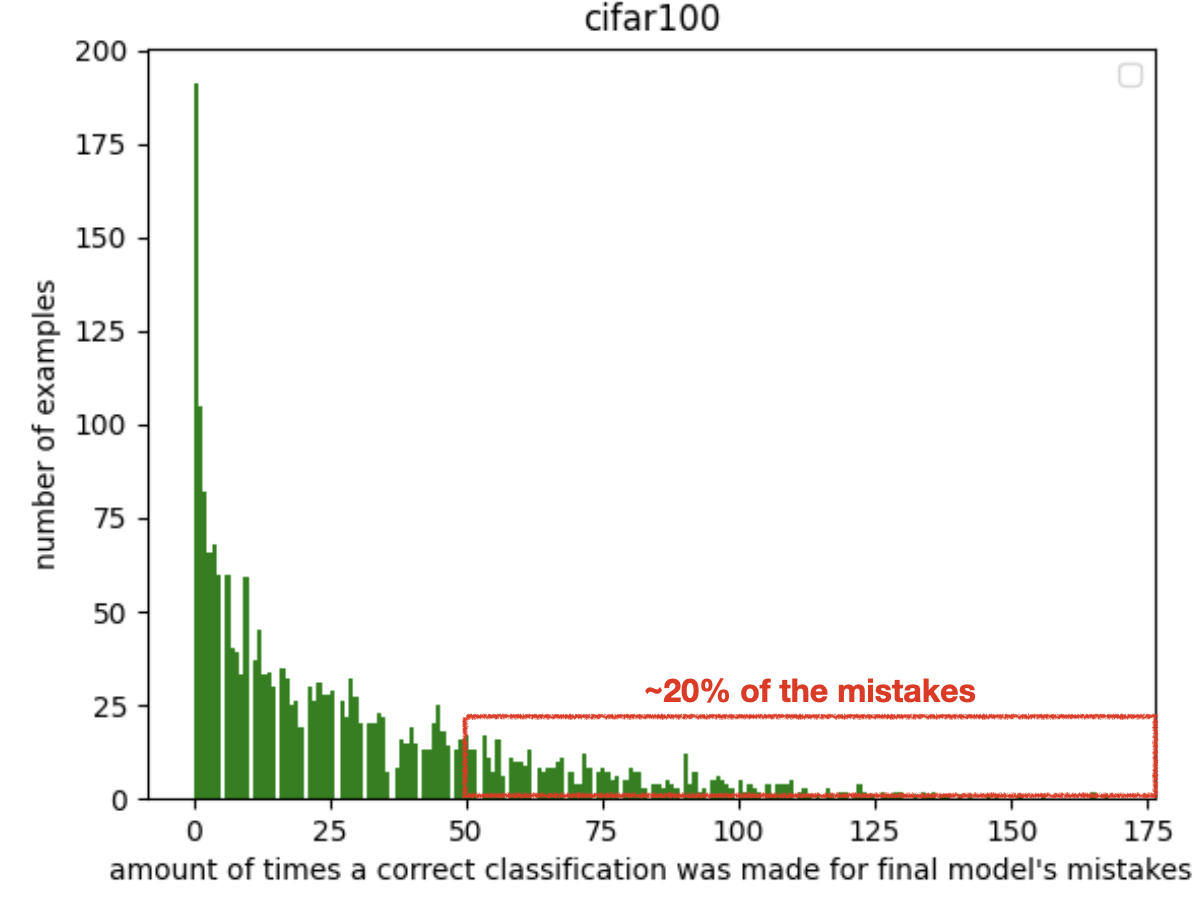}
    \vspace{-.5cm}
        \caption{cifar100}
    \end{subfigure}
    \begin{subfigure}[b]{0.24\linewidth}
        \includegraphics[width=\linewidth]{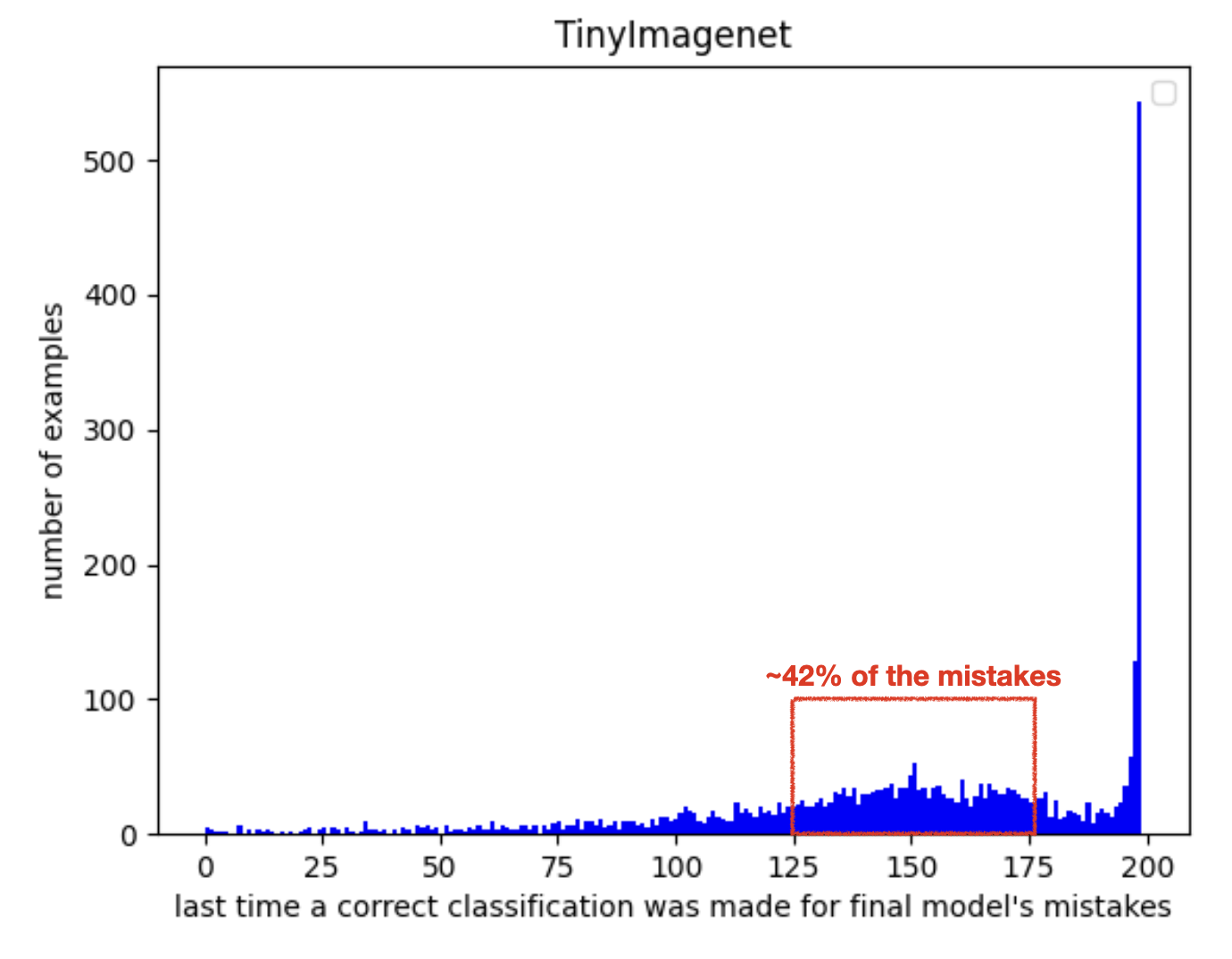}
    \vspace{-.5cm}
        \caption{TinyImagenet}
    \end{subfigure}
    \begin{subfigure}[b]{0.24\linewidth}
        \includegraphics[width=\linewidth]{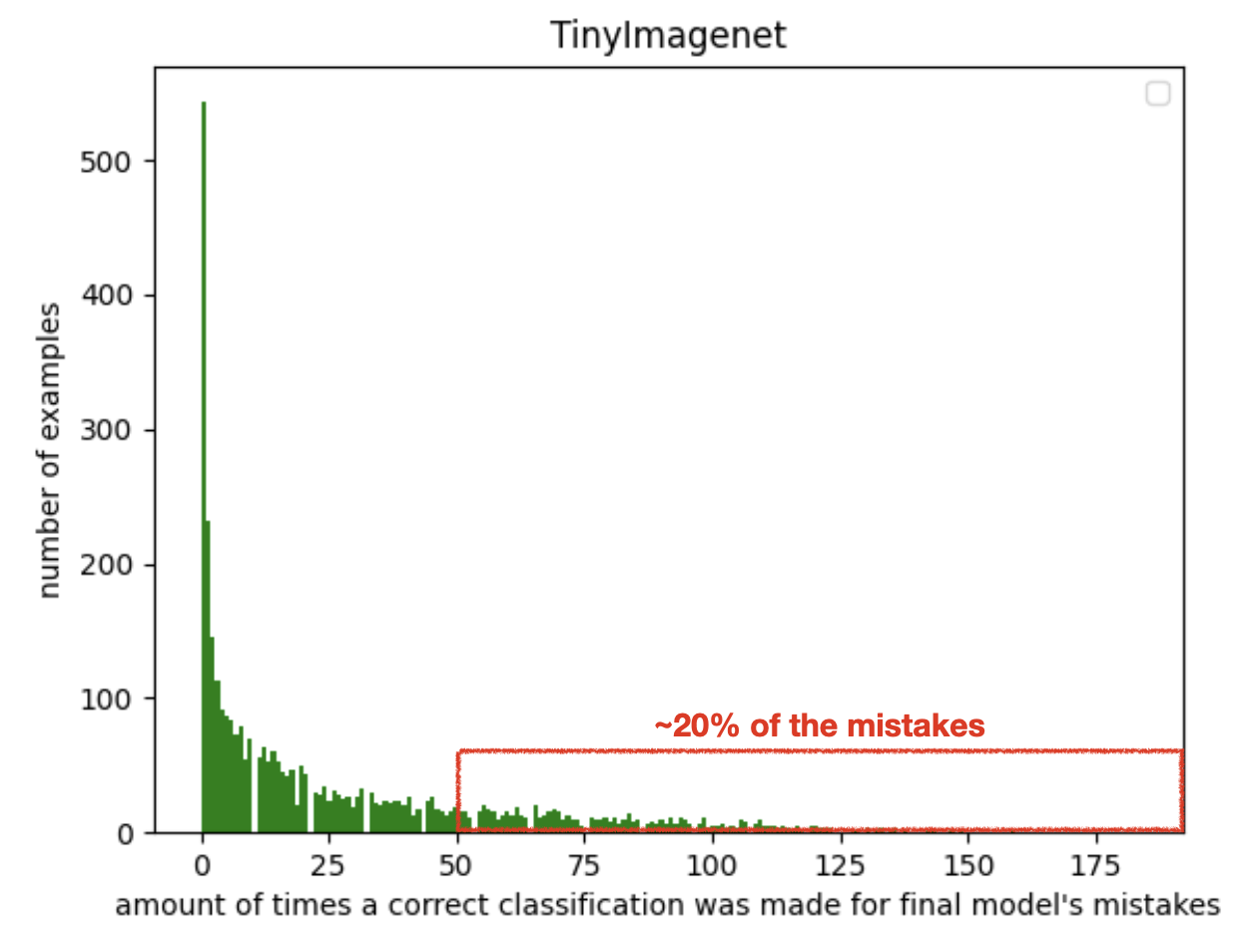}
    \vspace{-.5cm}
        \caption{TinyImagenet}
    \end{subfigure}
     \caption[ensemble size]{On the left: histogram of the new forget definition suggested by the reviewers - the last epoch in which a test data example mis-classified post training is correctly classified. One can see that within a span of 50 epochs a large fraction of those data examples are being "forgotten", as opposed to a random "forgetting" time which might have been expected. On the right: histogram of the amount of epochs a correct prediction was made for test data mis-classified post training; One can see that a large fraction of the mistakes were correctly predicted for over 25\% of the training, showing this correct prediction did not happen by random. Notably, both datasets are without label noise, and show no signs of overfit in the validation accuracy.} 
     \label{fig:differentforget}
     \vspace{-1.0em}
\end{figure*}

In this appendix, we show more examples of various neural network trained on different datasets which show significant forgetting during training (Fig~\ref{fig:moreforgetrate}), which demonstrates the generality of this phenomenon. We also show in Fig~\ref{fig:differentforget} that (i) many "forgotten" examples were correctly classified in a significant amount of epochs, indicating that their correct classification somewhere mid-training was not random, and (ii) that when observing the last epoch in which a correct classification was made for an example, one can see a dense span of epochs in which many examples where correctly predicted for the last time, which further supports that some subset of the data population was "forgotten" by the network in this span of epochs.

\section{Pseudo-code for our method}
\label{app:valalg}

In Alg~\ref{alg:test} and Alg~\ref{alg:val} we show how to implement our method and how to calculate its hyper-parameters, respectively. In the pseudo-codes we call functions that (i) calculate the probability for each class for a given example and a list of predictors (get\_class\_probabilities) (ii) calculate the forget value per epoch on some validation data, given the predictions at each epoch (calc\_forget\_per\_epoch) (iii) calculate the validation accuracy, to determine the best weights for each epoch (validation\_acc).

\begin{algorithm}[h]
   \caption{Knowledge Fusion (KF)}
\begin{algorithmic}
   \STATE {\bfseries Input:}{A set of checkpoints during training of the neural network \{$n_0$,...,$n_E$\}, w, test-point $x$
    \STATE{\bfseries Output:}  prediction for $x$}
    \STATE{\{$A_1$,...,$A_k$\},  \{$\epsilon_1,...,\epsilon_k$\} $\gets$ \textbf{calc\_early\_epochs\_and\_epsilons}(\{$n_0$,...,$n_E$\}) \hfill\textcolor{gray}{\#use Alg.\ref{alg:val}}}\; 
    \STATE{$class\_prob\_per\_checkpoint \gets$ 
    \STATE{\textbf{get\_class\_probabilities}(\{$n_0$,...,$n_E$\}, $x$)}}\;
    \STATE{$prob \gets class\_prob\_per\_checkpoint[E]$}\; 
    \FOR{$i\gets1$ {\bfseries to} $k$}
    \STATE{$prob_A \gets \mathbf{mean}(class\_prob\_per\_checkpoint[A_i-w:A_i+w])$}\;
    \STATE{$prob \gets \epsilon_i*prob_A+(1-\epsilon_i)*prob$}\;
    \ENDFOR
    \STATE{$prediction \gets \mathbf{argmax}(prob)$}\; 

    \STATE{\textbf{Return} $prediction$}
\end{algorithmic}
\label{alg:test}
\end{algorithm}
\begin{algorithm}[tb]
   \caption{KF - hyper-parameter calculation}
\begin{algorithmic}
   \STATE {\bfseries Input:}{all past checkpoints during training of the neural network \{$n_0$,...,$n_E$\}, w and validation data $V$, w and validation data $V$
    \STATE{\bfseries Output:}  list of alternative epochs and their weights}
   \STATE $class\_prob\_per\_checkpoint \gets$ 
\textbf{get\_class\_probabilities}(\{$n_0$,...,$n_E$\}, $V$)\;
    \STATE {$prob \gets class\_prob\_per\_checkpoint[E]$}\;
    \STATE {explore = \{$n_0$,...,$n_E$\}}\;
    \STATE {Alternative\_epochs = \{\}}\;
    \STATE {epsilons = \{\}}\;
   \WHILE{explore is not empty}
   \STATE {$F$ = \textbf{calc\_forget\_per\_epoch}($prob$, $class\_prob\_per\_checkpoint$)}\;
    \STATE {$alt\_epoch$ = \textbf{argmax}($F[explore]$)}\;    \STATE {Alternative\_epochs.\textbf{append($alt\_epoch$)}}\;
    \STATE {explore.\textbf{remove($alt\_epoch - 1$, $alt\_epoch$, $alt\_epoch + 1$)}}\;
    \FOR{$epsilon \in \{0, 0.01, ... , 1\}$}
    \STATE {$prob_A \gets \mathbf{mean}(class\_prob\_per\_checkpoint[A_i-w:A_i+w])$}\;
    \STATE {$combined\_prob \gets \epsilon*prob_A+(1-\epsilon)*prob$}\;
    \IF{\textbf{validation\_acc}($combined\_prob$) >= \textbf{validation\_acc}($prob$)}
    \STATE {$best\_prob = combined\_prob$}\;
    \STATE {$best\_epsilon = combined\_prob$}\;
    \ENDIF
    \ENDFOR
    \STATE {$prob = best\_prob$}\;
    \STATE {epsilons.append(\textbf{argmax}($best\_epsilon$))}\;
   \ENDWHILE
    \STATE{\textbf{Return} Alternative\_epochs, epsilons};
\end{algorithmic}
\label{alg:val}
\end{algorithm}

\section{Full implementation details}
\label{app:implementationdetails}
In our experiments we use three image classification datasets - Imagenet \citep{deng2009imagenet}, cifar100 \citep{krizhevsky2009learning} and TinyImagenet \citep{le2015tiny}. In the Imagenet experiments, we train all networks (Resnet \citep{he2016deep}, ConvNeXt \citep{liu2022convnet}, ViT \citep{dosoViTskiy2020image} and MaxViT \citep{tu2022MaxViT}) using the torchvision code\footnote{https://github.com/pytorch/vision/tree/main/references/classification} for training \citep{torchvision2016} with the recommended hyper-parameters, except ConvNeXt, which was trained using the official code\footnote{https://github.com/facebookresearch/ConvNeXt} with the recommended hyper-parameters, without using exponential moving average (EMA) (see App~\ref{abl:ema} for comparison to using EMA). With cifar100 and TinyImagenet, we train all networks for 200 epochs. For the clean versions of cifar100 and TinyImagenet we use batch-size of 32, learning rate of 0.01, SGD optimizer with momentum of 0.9 and weight decay of 5e-4, cosine annealing scheduler, and standard augmentations (horizontal flip, random crops). 

We use similar settings for our transfer learning experiments, in which the images are resized to $224 \times 224$, the learning rate is set to 0.001 and the network is initialized using Imagenet weights. In training, either the whole network is finetuned or only a new head (instead of the original fully connected layer), which consists of two dense layers, the first with output size of 100 times the embedding size. 

For noisy labels experiments, we train using cosine annealing with warm restarts (restarting the learning rate every 40 epochs), using a larger learning rate of 0.1 and updating it after every batch. We also use a larger batch size of 64 in cifar100 and 128 in TinyImagenet. In the suboptimal training described in Section~\ref{abl:subopttrain}, each image was cut before training into its central 224 over 224 pixels (images smaller than this size were first resized such that the smallest dimension was of size 224, then cut into 224 over 224). 

To obtain a fair comparison, we train the competitive methods in Table~\ref{table:specialmethods} from scratch using our network architecture and data. For \citep{huang2017snapshot} we train as instructed by the paper, while for \citep{izmailov2018averaging, garipov2018loss} we use our training scheme (as these methods are meant to be added to an existing training scheme) and performed hyper-parameter search to optimize the methods' performance in the new setting. Experiments were conducted on a cluster of GPU type A5000.

\paragraph{Injecting label noise} For the label noise experiments we inject noisy labels using two standard methods \cite{patrini2017making}:
\begin{enumerate}
[leftmargin=0.65cm,noitemsep]
\item \textbf{Symmetric noise:} a fraction $p \in \{0.2, 0.4, 0.6\}$ of the labels is selected randomly. Each selected label is switched to any other label with equal probability. 
\item \textbf{Asymmetric noise:} a fraction $p$ of the labels is selected randomly. Each selected label is switched to another label using a deterministic permutation function. 
\end{enumerate}
\section{Additional evaluations}
\label{app:additionaleval} 
We show here additional evaluation settings of our method and baselines on dataset with injected label noise, see Table~\ref{table:additional results}.

\begin{table}[thb!]
\footnotesize
  \centering
  \begin{tabular}{l| c||c|c}
    \multicolumn{1}{ c |}{Method/\textbf{Dataset}} & \multicolumn{1}{ c ||}{\textbf{CIFAR100 sym}} & \multicolumn{2}{ c }{\textbf{TinyImagenet}} 
    \\ 
    \multicolumn{1}{ r |}{\% label noise} &     60\% & 20\% & 40\% \\
    \toprule
        \emph{FGE}   & $38.3 \pm .7$ & $53.8 \pm .1$&$40.4 \pm .3$ \\
        \emph{SWA}   & $30.5 \pm .7$ & $52.5 \pm .2$&$39.4 \pm .3$ \\
        \emph{snapshot}   & $55.6 \pm .2$ & $62.6 \pm .1$ & $56.5 \pm .3$ \\
        \emph{KF (ours)}   & $\mathbf{57.6 \pm .2}$ & $\mathbf{62.8 \pm .2}$ &$\mathbf{57.0 \pm .5}$\\
  \end{tabular}
  \vspace{0.2cm}
  \caption{Mean (over random validation/test split) test accuracy (in percent) and standard error on image classification datasets with injected label noise, comparing our method and baselines.} 
  \label{table:additional results}
\end{table}

\section{Additional ablation results}
\label{app:additionalablation}
\subsection{Comparisons to additional baselines}
\label{abl:ema}
an alternative method to combine different checkpoints is to perform exponential moving average (EMA) during training, which is known to have some advantages \citep{polyak1992acceleration} and is used sometimes to reduce overfit \citep{liu2022convnet,dosoViTskiy2020image}, see \citep{tu2022MaxViT} for example). In table~\ref{table:ema} we explore this option for two datasets and a regular Resnet18, showing that our method can be of use when EMA doesn't work, or improves the performance much less than our method. 

\begin{table}[thb!]
\footnotesize
  \centering
  \begin{tabular}{l| c||c}
    \multicolumn{1}{ c |}{Method/\textbf{Dataset}} & \multicolumn{1}{ c ||}{\textbf{CIFAR-100}} & \multicolumn{1}{ c }{\textbf{TinyImagenet}} 
    \\ 
    \hline
        \emph{EMA}\scriptsize{ (decay = 0.999)}   & $-0.34 \pm .14$ & $0.73 \pm .11$ \\
        \emph{EMA }\scriptsize{ (decay = 0.9999)}   & $-0.06 \pm .33$ & $2.51 \pm .01$ \\
        \emph{KF}   & $\mathbf{1.05} \pm .14$ & $\mathbf{3.54} \pm .14$\\
  \end{tabular}
  \vspace{0.2cm}
  \caption{Mean (over random validation/test split) improvement in test accuracy (in percent) and standard error on image classification datasets, comparing our method and EMA with different decay values. We use the best epoch for EMA, calculated using the validation set.} 
  \label{table:ema}
\end{table}

\begin{table*}[thb!]
\footnotesize
  \centering
  \begin{tabular}{l| c || c|c|c || c|c || c}
    \multicolumn{1}{ c |}{Method/\textbf{Dataset}} & \multicolumn{1}{ c ||}{\textbf{CIFAR-100}} & \multicolumn{3}{ c ||}{\textbf{CIFAR-100 asym}} & \multicolumn{2}{ c || }{\textbf{CIFAR-100 sym}} & \multicolumn{1}{ c }{\textbf{TinyImagenet}} 
    \\ 
    \hline
    \multicolumn{1}{ r |}{\% label noise} &    0\% & 10\% & 20\% & 40\%  &  20\% & 40\% &  0\% \\
    \toprule
    \emph{ES}   & $78.13 \pm .4$ & $71.91 \pm .2$ & $68.54 \pm .3$ & $51.53 \pm .3$ & $68.16 \pm .4$ &  $61.17 \pm .2$ & $65.44 \pm .3$ \\
    \emph{TTA}  & $79.21 \pm .3$ & $72.97 \pm .1$ & $71.00 \pm .1$ & $54.53 \pm .1$ & $70.14 \pm .2$ & $63.59 \pm .1$ & $65.67 \pm .3$ \\
    \emph{ES + TTA}  & $79.11 \pm .2$ & $73.14 \pm .2$ & $70.46 \pm .2$ & $53.93 \pm .5$ & $70.21 \pm .1$ &  $63.57 \pm .1$ & $65.74 \pm .2$ \\
    \emph{KF (ours)}  & $78.71 \pm .2$ & $73.61 \pm .1$ & $71.24 \pm .5$ & $56.19 \pm .8$ & $72.21 \pm .3$ &  $65.75 \pm .1$ & $69.00 \pm .1$ \\
    \emph{KF (ours) + TTA}  & $\mathbf{79.55  \pm .3}$ & $\mathbf{74.83 \pm .1}$ & $\mathbf{72.65 \pm .2}$ & $\mathbf{57.71 \pm .3}$ & $\mathbf{72.33 \pm .2}$ &  $\mathbf{ 65.71 \pm .1}$& $\mathbf{69.05 \pm .3}$ \\

  \end{tabular}
  \vspace{0.2cm}
  \caption{Mean test accuracy (in percent) and standard error of resnet 18, comparing our method with Early Stopping (ES) and Test Time Augmentation (TTA) on datasets with and without label noise.} 

  \label{table:tta}
\end{table*}

\begin{table*}[thb!]
\scriptsize
  \centering
  \begin{tabular}{l| c|c|c||c|c|c}
    \multicolumn{1}{ c |}{Dataset/\textbf{Method}} & \multicolumn{3}{ c ||}{\textbf{original model}} & \multicolumn{3}{ c }{\textbf{KF}} 
    \\ 
        \hline
    \multicolumn{1}{ r |}{evaluation metric} & natural accuracy&  transformed accuracy & bias & natural accuracy& transformed accuracy & bias \\
    \hline
        \emph{cifar10 w/o color}   & $89.07 \pm .48$ & $87.98 \pm .38$ & $0.07 \pm .001$ & $89.90 \pm .40$ & $87.85 \pm .48$ & $0.07 \pm .002$ \\
        \emph{cifar10 center cropped to 28x28}   & $88.45 \pm .31$ & $70.44 \pm .44$ & $0.13 \pm .003$ & $88.92 \pm .32$ & $70.21 \pm .74$ & $0.13 \pm .004$\\
        \emph{cifar10 downsampled to 16x16}   & $85.43 \pm .32$ & $76.70 \pm .13$ & $0.08  \pm .001$ & $86.55 \pm .27$ & $77.47 \pm 14$ & $0.07  \pm .001$\\
        \emph{cifar10 downsampled to 8x8}   & $80.061 \pm .33$ & $52.03 \pm .49$ & $0.22 \pm .002$ & $81.48 \pm .44$ & $52.99 \pm .49$ & $0.21 \pm .003$\\
        \emph{cifar10 with Imagenet replacements}   & $88.45 \pm .31$ & $70.44  \pm .44$ & $0.13 \pm .003$ & $88.92 \pm .32$ & $70.44 \pm .44$ & $0.13 \pm .004$\\
  \end{tabular}
  \vspace{0.2cm}
  \caption{Mean (over random validation/test split) test accuracy and amplification bias (in percent) and standard error on natural and transformed test sets, comparing our method and the original model.} 
  \label{table:fairness}
\end{table*}

Our analysis focused, for the most part, on overfit that exists even in the scenario in which test accuracy does not decrease as training proceeds - which means that "early stopping" - culminating the training when performance over validation data decreases - has a minimal effect. Still, we wanted to compare our performance to early stopping (ES) on datasets with and without label noise. We also included in this comparison the test-time augmentation (TTA) method, in which a test example is being classified several times, each time with a different augmentation, and given a final classification based on the average class probabilities of the different classifications. The results in Table~\ref{table:tta} indicate that our method is comparable or better than both methods even when label noise exists in the training dataset (which leads to deteriorating performance as training proceeds), and that it is complementary to test-time augmentation.


\subsection{model size}
\label{subsec:modelsize}

\begin{table}[thb!]
\scriptsize
  \centering
  \begin{tabular}{l| c||c||c}
    \multicolumn{1}{ c |}{Method/\textbf{model size}} & \multicolumn{1}{ c ||}{\textbf{small}} & \multicolumn{1}{ c || }{\textbf{base}} & \multicolumn{1}{ c }{\textbf{large}} 
    \\ 
    \hline
        \emph{single network}   & $83.21 \pm .01$ & $83.31 \pm .15$ & $82.92 \pm .09$ \\
        \emph{KF (ours)}   & $83.17 \pm .04$ & $\mathbf{83.57 \pm .15}$ &$\mathbf{83.96 \pm .09}$\\
  \end{tabular}
  \vspace{0.2cm}
  \caption{Mean (over random validation/test split) test accuracy (in percent) and standard error on image classification datasets, comparing our method and the original predictor (ConvNeXt, trained on Imagenet) with varying number of parameters} 

  \label{table:modelsize}
\end{table}

A common practice nowadays is to use very large neural networks, with hundred of millions parameters, or even more. However, enlarging models does not always improve performance, as large number of parameters can lead to overfit. In Fig.~\ref{subfig:modelsizeforget} we show that indeed larger versions of a model can cause increasing forget fraction, which also improves the benefit of our model (see Table~\ref{table:modelsize}), making it especially useful when one uses a large model.


\subsection{Comparison to a regular ensemble}

\label{subsec:regens}
\begin{figure}[h]
    \centering
    \begin{subfigure}[b]{0.45\linewidth}
        \includegraphics[width=\linewidth]{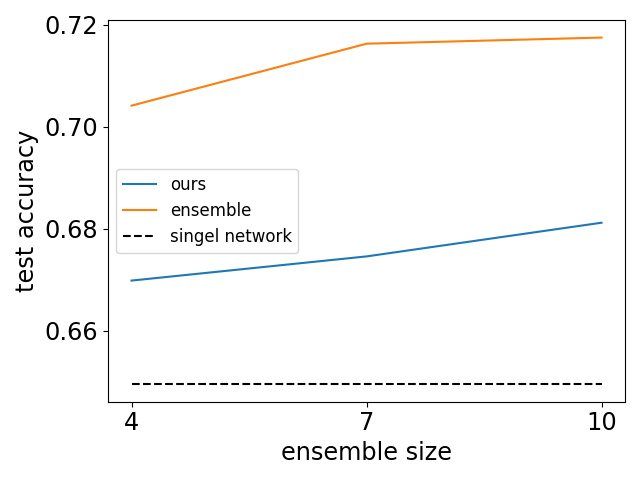}
    \vspace{-.5cm}
        \caption{TinyImagenet}
        \label{subfig:oursvsensembleTinyImg}
    \end{subfigure}
    \begin{subfigure}[b]{0.45\linewidth}
        \includegraphics[width=\linewidth]{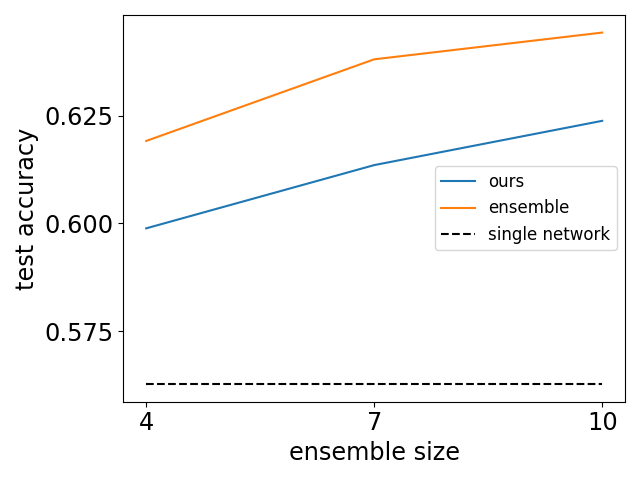}
    \vspace{-.5cm}
        \caption{TinyImagenet, 20\% sym noise}
        \label{subfig:oursvsensembleTinyImg20lb}
    \end{subfigure}
    \begin{subfigure}[b]{0.45\linewidth}
        \includegraphics[width=\linewidth]{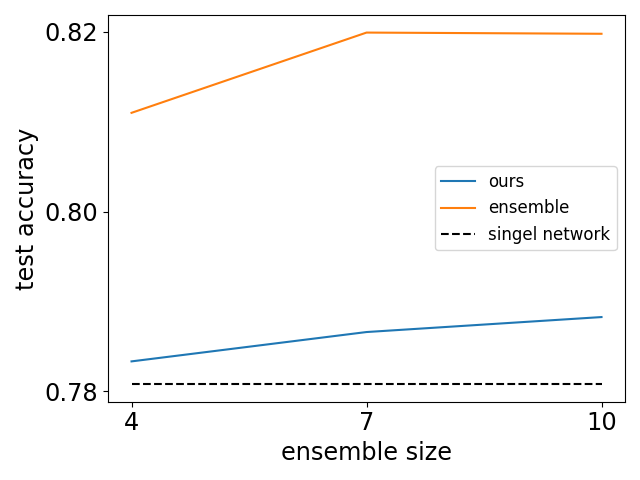}
    \vspace{-.5cm}
        \caption{CIFAR100}
        \label{subfig:oursvsensembleC100}
    \end{subfigure}
    \begin{subfigure}[b]{0.45\linewidth}
        \includegraphics[width=\linewidth]{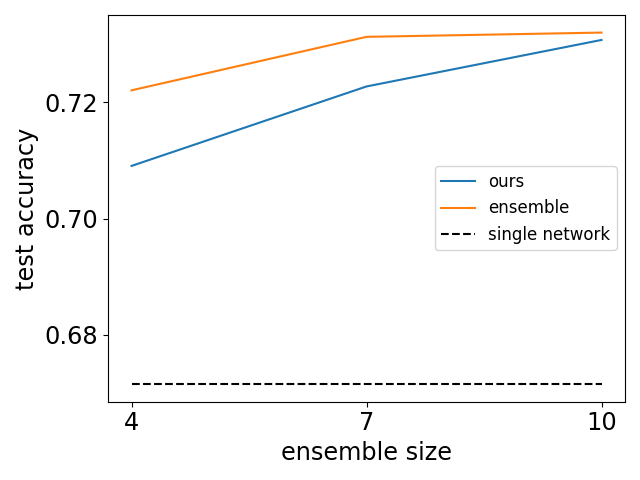}
    \vspace{-.5cm}
        \caption{CIFAR100,  20\% asym noise}
        \label{subfig:oursvsensembleC100asym20}
    \end{subfigure}
     \caption[forgetrate]{Comparing our method with limited number of checkpoints and an ensemble of the same size of independent networks} 
     \label{fig:oursvsensemble}
\end{figure}

A regular ensemble, unlike ours, requires multiple training of independent networks, which could be unfeasible. Thus, it serves as an "upper bound" for our method's performance. In Fig.~\ref{fig:oursvsensemble}, we compare our method and a regular ensemble of the same size, showing our method can achieve much of the performance gain provided by the regular ensemble. Notably, when label noise occurs our method can add most, if not all, of the regular ensemble performance gain.

\subsection{Fairness}
\label{abl:fairness}
In this section we study our method's effect on the model's \emph{fairness}, i.e. the effect non-relevant features have on the classification of test data examples. We follow \citep{wang2020towards} and train and test our models on datasets in which they might learn spurious correlations. To create those datasets, we divide the classes into two groups: in each class of the first group 95\% of the training images goes through a transformation (and the rest remain unchanged), and vice versa for the classes of the second group. The transformation we use are: removing color, lowering the images resolution (by down sampling and up sampling), and replacing images with downsampled images for the same class in Imagenet. We use cifar10 in our evaluation as done in  \citep{wang2020towards}, and use also cifar100 with the remove color transformation (the rest of the transformations were less appropriate for this datasets, as it contains similar classes that could actually become harder to seperate at a lower resolution). We use the same method as before for our validation data, and thus the validation is of the same distribution as the test data.

Our evaluation use the following metrics: (i) the test accuracy on two test sets (with/out the transformation), which should be lower if the model learns more spurious correletions, and (ii) the amplification bias defined in \citep{zhao2017men}, which is defined is follows:

\begin{equation}
    \frac{1}{|C|}\sum_{c \in C} \frac{max(c_T, c_N)}{c_T+c_N} - 0.5
\end{equation}

When $C$ is the group of classes, $c_T$ is the number of images from the transformed test set predicted to be of class c, and $c_N$ is the number of images from the natural test set predicted to be of class c - we would like those to be as close as possible, since the transformation shouldn't change the prediction, and thus the lower the score the better. 

The results of our evaluation are presented in Table~\ref{table:fairness}. To summarize, our method improves the average performance on both datasets without deteriorating the amplification bias, which indicates that our method has no negative effects on the model's fairness.

\end{document}